\documentclass{llncs}

\usepackage[parfill]{parskip}    % Activate to begin paragraphs with an empty line rather than an indent
\usepackage[version=0.96]{pgf}
\usepackage{tikz}
\usetikzlibrary{arrows,matrix,graphs,shapes,snakes,automata,backgrounds,petri,positioning,calc,decorations.pathmorphing}
\usepackage[latin1]{inputenc}
\usepackage{verbatim}
\usepackage[leqno]{amsmath}
\usepackage{bm}
\usepackage{amssymb}
\usepackage{latexsym}
\usepackage{proof}
\usepackage{array}
\usepackage{calc}
\usepackage{mathdots}
\usepackage{stmaryrd}
\usepackage{harvard}
\usepackage[english]{babel}
\usepackage{extarrows}
\usepackage{rotating}
\usepackage{hyperref}
\usepackage{linguex}

\newcommand{\splitc}{\,\check{\,}}
\newcommand{\bridge}{\,\hat{\,}}
\newcommand{\smallplus}{\scalebox{.5}{+}}
\newcommand{\smallminus}{\scalebox{.5}{\ensuremath{-}}}

\newcommand{\downl}{\downarrow_{>}}
\newcommand{\downr}{\downarrow_{<}}
\newcommand{\upl}{\uparrow_{>}}
\newcommand{\upr}{\uparrow_{<}}
\newcommand{\prodl}{\odot_{>}}
\newcommand{\prodr}{\odot_{<}}

\tikzstyle{reverseclip}=[insert path={(current page.north east) --
  (current page.south east) --
  (current page.south west) --
  (current page.north west) --
  (current page.north east)}
]

\newcommand{\rmindices}{
\node at (8.25em,3.6em) {$\overbrace{\qquad\qquad\qquad\qquad\qquad\qquad\qquad\qquad\ }^{C}$};
\node at (0em,2.5em) {$\scriptstyle x_0$};
\node at (7.0em,2.5em) {$\scriptstyle x_n$};
\node at (14.75em,2.5em) {$\scriptstyle x_{n\smallplus m\smallminus 1}$};
\node at (17.25em,2.5em) {$\scriptstyle x_{n\smallplus m}$};}
\newcommand{\rmindicesp}{
\node at (8.25em,3.6em)
{$\overbrace{\qquad\qquad\qquad\qquad\qquad\qquad\qquad\qquad\
  }^{A\prodr B}$};
\node at (0em,2.5em) {$\scriptstyle x_0$};
\node at (7.0em,2.5em) {$\scriptstyle x_n$};
\node at (14.75em,2.5em) {$\scriptstyle x_{n\smallplus m\smallminus 1}$};
\node at (17.25em,2.5em) {$\scriptstyle x_{n\smallplus m}$};}
\newcommand{\rmindicess}{
\node at (8.25em,3.6em) {$\overbrace{\qquad\qquad\qquad\qquad\qquad\qquad\qquad\qquad\
  }^{tus}$};
\node at (0em,2.5em) {$\scriptstyle x_0$};
\node at (7.0em,2.5em) {$\scriptstyle x_n$};
\node at (14.75em,2.5em) {$\scriptstyle x_{n\smallplus m\smallminus 1}$};
\node at (17.25em,2.5em) {$\scriptstyle x_{n\smallplus m}$};}

\newcommand{\editout}[1]{}

\newlength{\bre}
\newcolumntype{C}{@{}>{\rule{0mm}{\bre}}p{\bre}<{}@{}}

\title{Extended Lambek calculi \\ and first-order linear logic}
\author{Richard Moot}
\institute{CNRS, LaBRI, Unversity of Bordeaux\\
\email{Richard.Moot@labri.fr}}

\begin{document}

\maketitle

\section{Introduction}

The Syntactic Calculus \cite{lambek} --- often simply called the Lambek calculus, L,  --- is a beautiful system in many ways: Lambek grammars
give a satisfactory syntactic analysis for the (context-free) core of natural language
and, in addition, it
provides a simple and elegant syntax-semantics interface.

However, since Lambek grammars generate only context-free languages \cite{pentus97},
there are some well-know linguistic phenomena (Dutch verb clusters, at least if
we want to get the semantics right \cite{weakin}, Swiss-German verb clusters \cite{shieber}, etc.)
which cannot be treated by Lambek grammars. 

%The class of mildly
%context-sensitive languages \cite{mcs} is an interesting family of language classes extending  the context-free languages in a limited way. Though there is no %agreement as to which exact member of this family is most suited for the description of natural language, there seems to be a consensus that we need a %formalism able to generate at least the tree adjoining languages, and possibly slightly more. 

In addition, though the syntax-semantics interface works for many of
the standard examples, the Lambek calculus does not allow
a non-peripheral quantifier to take wide scope (as we would need for sentence~\ref{ex:medq} below if we want the existential quantifier to have wide scope, the so-called ``de re'' reading) or non-peripheral extraction (as illustrated by sentence~\ref{ex:medex} below); see \cite[Section~2.3]{M10} for discussion.

\ex. \label{ex:medq} John believes someone left.
%\ex. 

\ex. \label{ex:medex} John picked up the package which Mary left yesterday.
%\end{enumerate}

To deal with these problems, several extensions of the Lambek calculus have been proposed. Though this is not the time and place to review them --- I recommend \cite{m10sep,M10} and the references cited therein for an up-to-date overview of the most prominent extensions; they include multimodal categorial grammar \citeaffixed{mmli}{MMCG,}, the Lambek-Grishin calculus \citeaffixed{moortgat07sym}{LG,} and the Displacement calculus \citeaffixed{mv11displacement}{D,} --- I will begin by listing a number of properties which I consider desirable for such an extension. In essence, these desiderata are all ways of keeping as many of good points of the Lambek calculus as possible while at the same time dealing with the inadequacies sketched above.\footnote{To the reader who is justifiably skeptical of any author who writes down a list of desiderata, followed by an argument by this same author arguing how well he scores on his own list, I say only that, in my opinion, this list is uncontroversial and at least implicitly shared by most of the work on extensions of the Lambek calculus and that the list still allows for a considerable debate as to \emph{how well} each extension responds to each desideratum as well as discussion about the relative importance of the different items.}

\begin{enumerate}
\item simple proof theory, 
\item generate the mildly context-sensitive languages,
\item simple syntax-semantics interface giving a correct and simple account of medial scope for quantifiers
  and of medial extraction,
\item have a reasonable computational complexity.
\end{enumerate}

None of these desiderata is absolute: there are matters of degree for
each of them. First of all, it is often hard to distinguish familiarity
from simplicity, but I think that having multiple equivalent proof
systems for a single calculus is a sign that the calculus is a natural
one: the Lambek calculus has a sequent calculus, natural deduction,
proof nets, etc.\ and we would like its extensions to have as many of
these as possible, each formulated in the simplest possible way. 

The mildly context-sensitive languages \cite{mcs} are a family of
languages which extend the context-free language in a limited way, and opinions vary as to which of the members of this
family is the most appropriate for the description of natural language. Throughout this article, I will only make the (rather conservative and uncontroversial) claim that any extension of the Lambek calculus should at least generate the tree adjoining languages, the multiple context-free languages \cite{mcfg} (the well-nested MCFLs \cite{kanazwaw09wn} are weakly equivalent to the tree adjoining languages) or the simple, positive range concatenation grammars \citeaffixed{boullier}{sRCG, weakly equivalent to MCFG,}.

With respect to the semantics, it generally takes the form of a simple homomorphism from proofs in the source logic to proofs in the Lambek-van Benthem calculus LP (which is multiplicative intuitionistic linear logic, MILL, for the linear logicians), though somewhat more elaborate continuation-based mappings \cite{bernardi07conti} have been used as well.

%Moortgat (find citation): tension between
%LP (for semantics) and NL (for syntax). Simple homomorphism to proofs
%in MLL/LP.

Finally, what counts as reasonable
computational complexity is open to discussion as well: since
theorem-proving for the Lambek calculus is NP complete \cite{pentus06np}, I will
consider NP-complete to be ``reasonable'', though polynomial parsing is
generally considered a requirement for mildly context-sensitive
formalisms \cite{mcs}. Since the complexity of the logic used corresponds to the \emph{universal} recognition problem in formal language theory, NP completeness is not as bad as it may seem, since it corresponds to the complexity of the universal recognition problem for multiple context-free grammars (when we fix the maximum number of string tuples a non-terminal is allowed to have), which is a prototypical mildly context-sensitive formalism. Little is known on polynomial \emph{fixed} recognition for extended Lambek calculi (though some partial results can be found in \cite{diss,moot08tag}). The fixed recognition problem for the Lambek calculus itself is known to be polynomial \cite{polyorder}.

\editout{
In general, there are two basic families of extensions to the Lambek
calculus: 

\begin{itemize}
\item the first family replaces the strings, which are the
elementary units of the calculus by \emph{structures} --- (labeled) trees ---
and specify logical rules for the reconfiguration of these
trees. Examples of formalism in this family are multimodal categorial
grammar (MMCG) and
the Lambek-Grishin (LG) calculus. In a larger sense, it would also include
tree adjoining grammars and combinatory categorial grammars.
\item the second family uses \emph{tuples} of strings. Examples of
  frameworks in this family are the discontinuous Lambek calculus (D) and
  multiplicative first-order intuitionistic linear logic (MILL1), the two main
  protagonists of the current article. In a large sense, it would also
  include abstract categorial grammars, multiple context-free grammars, range concatenation grammars
  and tupled pregroup grammars.
\end{itemize}
}

\begin{table}

\begin{center}
\begin{tabular}{lllcc}
Calculus & Complexity & Languages & Scope & Extraction \\
L & NP complete & CFL & -- & -- \\
MMCG & PSPACE complete & CSL & + & + \\
LG & NP complete & $\geq$ MCFL & + & -- \\
D & NP complete & $\geq$ MCFL & + & + \\
%ACG$_2$ & NP complete & MCFL & -- & -- \\
%ACG$_3$ & EXPSPACE hard & $\geq$ MCFL & + & + \\
MILL1 & NP complete & $\geq$ MCFL & + & + \\
\end{tabular}
\end{center}
\vspace{.5\baselineskip}

\caption{The Lambek calculus and several of its variants/extensions, together with the complexity of the universal recognition problem, classes of languages generated and the appropriateness of the formalism for handling medial quantifier scope and medial extraction}
\label{tab:extensions}
\end{table}

Table~\ref{tab:extensions} gives an overview of the Lambek calculus as well as several of its prominent extensions with respect to the complexity of the universal recognition problem, the class of languages generated and the facilities in the formalism for handling medial quantifier scope and medial extraction.

In this paper, I will present an alternative extension of the Lambek calculus: first-order multiplicative intuitionistic linear logic (MILL1) \cite{quant,mill1}. It generates the right
class of languages (MCFG are a subset of the Horn clause fragment, as shown in Section~\ref{sec:mcfg}),
and embeds the simple Displacement calculus (D, as shown in Section~\ref{sec:trans} and \ref{sec:correct}). As can be seen in Table~\ref{tab:extensions}, it has the lowest complexity class among the different extensions, generates (at least) the right class of languages, but also handles medial scope and medial extraction in a very simple way (as shown already in \cite{mill1}). In addition, as we will see in Section~\ref{sec:mill1}, MILL1 has a very simple proof theory, essentially a resource-conscious version of first-order logic, with a proof net calculus which is a simple extension of the proof nets of multiplicative linear logic \cite{multiplicatives,quant}. Finally, the homomorphism from MILL1 to MILL for semantics consists simply of dropping the first-order quantifiers.

I will also look at the Displacement calculus from the perspective of MILL1 and give a translation of D into MILL1, indirectly solving
two open problems from \cite{morrill2010} by providing a proof
net calculus for D and showing that D is NP-complete. In addition it is also worth mentioning briefly that the simpler proof theory of MILL1 (ie.\ proof nets) makes proving cut elimination for D very easy: as for the multiplicative case, cut elimination for MILL1 consists of simple, local conversions only with only three distinct cases to verify (axiom, tensor/par and existential/universal).

The remainder of this paper is structured as follows. In the next section, I will briefly introduce MILL1 and its proof theory, including a novel correctness condition for first-order proof nets, which is a simple extension of the contraction criterion from \citeasnoun{reductions}. Section~\ref{sec:d} will introduce the Displacement calculus, D, using a presentation of the calculus from \cite{mv11displacement} which emphasizes the operations on string tuples and, equivalently, on string positions. Section~\ref{sec:trans} will present a translation from D to MILL1, with a correctness proof in Section~\ref{sec:correct}. Section~\ref{sec:other} will briefly mention some other possible applications of MILL1, which include agreement, non-associativity and island constraints and quantifier scope restrictions. Finally, I will reflect on the implications of the results in this paper, giving some interesting open problems.

\section{MILL1}
\label{sec:mill1}

First-order multiplicative intuitionistic linear logic (MILL1) extends (multiplicative) intuitionistic linear logic with the first-order quantifiers $\exists$ and $\forall$.
The first-order multiplicative fragment shares many of the good properties of the propositional fragment: the decision problem is NP complete  \cite{lincoln} and it has a simple proof net calculus which is an extension of the proof net calculus for multiplicative linear logic.

Table~\ref{tab:millnd} presents the natural deduction calculus for MILL1, which is without surprises, though readers familiar with intuitionistic logic should note that the $\otimes E$, $\multimap I$ and $\exists E$ rule discharge exactly one occurrence of each of the hypotheses with which it is coindexed.

\begin{table}
$$
\begin{array}{ccc}
\infer[\otimes E_i]{C}{A \otimes B & \infer*{C}{[A]^i[B]^i}}
&&
\infer[\otimes I]{A\otimes B}{A & B} \\
\\
\infer[\multimap E]{B}{A & A \multimap B} &&
\infer[\multimap I]{A \multimap B}{\infer*{B}{[A]^i}} \\
\\
\infer[\exists E_i^*]{C}{\exists x .A & \infer*{C}{[A]^i}} &&
\infer[\exists I]{\exists x.A}{A[x:=t]} \\
\\
\infer[\forall E]{A[x:=t]}{\forall x. A} & &
\infer[\forall I^*]{\forall x. A}{A} \\
\end{array}
$$

\vspace{.8ex}
\begin{center}
$\rule{0pt}{1ex}^*$ no free occurrences of $x$ in any of the free hypotheses
\end{center}
%(except the cancelled hypothesis $A$ for the $\exists E$ rule
\caption{Natural deduction rules for MILL1}
\label{tab:millnd}
\end{table}

I will present the proof net calculus in three steps, which also form a basic proof search procedure: for a given statement $\Gamma \vdash C$ (with $C$ a formula and $\Gamma$ a multiset of formulas) we form a \emph{proof frame} by unfolding the formulas according to the logical links shown in the bottom two rows of Table~\ref{tab:links}, using the negative unfolding for the formulas in $\Gamma$ and the positive unfolding for the formula $C$. We then connect the atomic formulas using the axiom link (shown on the top left of the table) until we have found a complete matching of the atomic formulas, forming a \emph{proof structure}. Finally, we check if the resulting proof structure is a \emph{proof net} (ie.\ we verify if $\Gamma\vdash C$ is derivable) by verifying it satisfies a correctness condition. 

As is usual, I will use the following conventions, which will make formulating the proof net calculus simpler.

\begin{itemize}
\item dotted binary links are called \emph{par} links, solid binary links are called \emph{tensor} links,
\item dotted unary links are called \emph{universal} links, solid unary links are called existential links, the bound variables of these links are called universally bound and existentially bound respectively.
\item each occurrence of a quantifier link uses a distinct bound variable,
\item the variable of a positive $\forall$ and a negative $\exists$ link (ie.\ the universal links and universally quantified variables) are called its \emph{eigenvariable},
\item following \cite{empires}, I require eigenvariables of existential links to be used \emph{strictly}, meaning that replacing the eigenvariable throughout a proof with a special, unused constant will not result in a proof (in other words, we never unnecessarily instantiate an existentially quantified variable with the eigenvariable of a universal link).
\end{itemize}

The fact that par links and universal links are drawn with dotted lines is not a notational accident: one of the fundamental insights of focusing proofs and ludics \cite{focus,locussolum} is that these two types of links naturally group together, as do the existential and tensor links, both drawn with solid lines. This property is also what makes the correctness proof of Section~\ref{sec:correct} work. When it is convenient to refer to the par and universal links together, I will call them \emph{asynchronous} links, similarly I will refer to the existential and tensor links as \emph{synchronous} links (following \citeasnoun{focus}).

\editout{
\begin{table}
$$
\begin{array}{ccc}
\infer[\textit{Ax}]{A \vdash A}{} & &
\infer[\textit{Cut}]{\Gamma, \Delta \vdash C}{\Gamma \vdash A & \Delta,A\vdash C} \\
\\
\infer[L\otimes]{\Gamma,A\otimes B \vdash C}{\Gamma, A, B \vdash C} &&
\infer[R\otimes]{\Gamma,\Delta \vdash A \otimes B}{\Gamma \vdash A & \Delta \vdash B}\\
\\
\infer[L\multimap]{\Gamma,\Delta,A\multimap B \vdash C}{\Delta \vdash A & \Gamma,B \vdash C} &&
\infer[R\multimap]{\Gamma \vdash A \multimap B}{\Gamma, A \vdash B} \\
\\
\infer[L\exists^*]{\Gamma,\exists x.A \vdash C}{\Gamma,A \vdash C} &&
\infer[R\exists]{\Gamma \vdash \exists x.C}{\Gamma \vdash C[x := t]} \\
\\
\infer[L\forall]{\Gamma, \forall x.A \vdash C}{\Gamma A[x := t] \vdash C} &&
\infer[R\forall^*]{\Gamma \vdash \forall x. A}{\Gamma \vdash A} \\
\end{array}
$$

\vspace{.8ex}
\begin{center}
$\rule{0pt}{1ex}^*$ no free occurrences of $x$ in $\Gamma, A, C$
\end{center}

\caption{Sequent calculus for MILL1}
\end{table}}

\begin{table}

\begin{center}
\begin{tikzpicture}
\node (anx) at (0em,0em) {$\overset{-}{A}$};
\node (anp) at (6em,0em) {$\overset{+}{A}$};
\draw (anx) -- (0em,2em) -- (6em,2em) -- (anp);
\node (cnx) at (12em,0em) {$\overset{-}{A}$};
\node (cnp) at (18em,0em) {$\overset{+}{A}$};
\draw (cnx) -- (12em,-2em) -- (18em,-2em) -- (cnp);
\end{tikzpicture}
\end{center}
\vspace{\baselineskip}
\begin{center}
\begin{tikzpicture}
\node (forallnc) {$\overset{-}{\forall x. A}$};
\node (forallnp) [above=2em of forallnc] {$\overset{-}{A[x:=t]}$};
\draw (forallnc) -- (forallnp);
\node (forallpc) [right=7em of forallnc] {$\overset{+}{\forall x. A}$};
\node (forallpp) [above=2em of forallpc] {$\overset{+}{A}$};
\draw [dotted] (forallpc) -- (forallpp);
\node (existsnc) [right=7em of forallpc] {$\overset{-}{\exists x. A}$};
\node (existsnp) [above=2em of existsnc] {$\overset{-}{A}$};
\draw [dotted] (existsnc) -- (existsnp);
\node (otimesnc) [above=7em of existsnc] {$\overset{-}{A\otimes B}$};
\node (tmponl) [left=0.66em of otimesnc] {};
\node (aotimesnc) [above=2.5em of tmponl] {$\overset{-}{A}$};
\node (tmponr) [right=0.66em of otimesnc] {};
\node (botimesnc) [above=2.5em of tmponr] {$\overset{-}{B}$};
\begin{scope}
\begin{pgfinterruptboundingbox}
\path [clip] (otimesnc.center) circle (2.5ex) [reverseclip];
\end{pgfinterruptboundingbox}
\draw [dotted] (otimesnc.center) -- (botimesnc);
\draw [dotted] (otimesnc.center) -- (aotimesnc);
\end{scope}
\begin{scope}
\path [clip] (aotimesnc) -- (otimesnc.center) -- (botimesnc);
\draw (otimesnc.center) circle (2.5ex);
\end{scope}
\node (otimespc) [right=7em of otimesnc] {$\overset{+}{A\otimes B}$};
\node (tmpopl) [left=0.66em of otimespc] {};
\node (aotimespc) [above=2.5em of tmpopl] {$\overset{+}{A}$};
\node (tmpopr) [right=0.66em of otimespc] {};
\node (botimespc) [above=2.5em of tmpopr] {$\overset{+}{B}$};
\draw (otimespc) -- (aotimespc);
\draw (otimespc) -- (botimespc);
\node (existspc) [below=7em of otimespc] {$\overset{+}{\exists x. A}$};
\node (existspp) [above=2em of existspc] {$\overset{+}{A[x:=t]}$};
\draw  (existspc) -- (existspp);
\node (lollinc) [above=7em of forallnc] {$\overset{-}{A\multimap B}$};
\node (tmplnl) [left=0.66em of lollinc] {};
\node (alollin) [above=2.5em of tmplnl] {$\overset{+}{A}$};
\draw (lollinc) -- (alollin);
\node (tmplnr) [right=0.66em of lollinc] {};
\node (blollin) [above=2.5em of tmplnr] {$\overset{-}{B}$};
\draw (lollinc) -- (blollin);
\node (lollipc) [above=7em of forallpc] {$\overset{+}{A\multimap B}$};
\node (tmplpl) [left=0.66em of lollipc] {};
\node (alollip) [above=2.5em of tmplpl] {$\overset{-}{A}$};
\node (tmplpr) [right=0.66em of lollipc] {};
\node (blollip) [above=2.5em of tmplpr] {$\overset{+}{B}$};
\begin{scope}
\begin{pgfinterruptboundingbox}
\path [clip] (lollipc.center) circle (2.5ex) [reverseclip];
\end{pgfinterruptboundingbox}
\draw [dotted] (lollipc.center) -- (blollip);
\draw [dotted] (lollipc.center) -- (alollip);
\end{scope}
\begin{scope}
\path [clip] (alollip) -- (lollipc.center) -- (blollip);
\draw (lollipc.center) circle (2.5ex);
\end{scope}
\end{tikzpicture}
\end{center}

\caption{Logical links for MILL1 proof structures}
\label{tab:links}
\end{table}

In Table~\ref{tab:links}, the formulas drawn below the link are its conclusions (the axiom link, on the top left of the table, is the only multiple conclusion link, the cut link, on the top right, does not have a conclusion, all logical links have a single conclusion), the formulas drawn above the link are its premisses.

\begin{definition}
%Let $\Gamma \vdash C$ be a proof with active hypotheses $\Gamma$ and conclusions $C$.
A \emph{proof structure} is a set of polarized formulas connected by instances of the links shown in Table~\ref{tab:links} such that each formula is at most once the premiss of a link and exactly once the conclusion of a link. Formulas which are not the premiss of any link are called the \emph{conclusions} of the proof structure. We say a proof structure with negative conclusions $\Gamma$ and positive conclusions $\Delta$ is a proof structure of the statement $\Gamma \vdash \Delta$.
\end{definition}

\begin{definition}
Given a proof structure $\Pi$ a \emph{switching} is
\begin{itemize}
\item for each of the par links a choice of one of its two premisses,
\item for each of the universal links a choice either of a formula containing the eigenvariable of the link or of the premiss of the link.
\end{itemize}
\end{definition}

\begin{definition}
Given a proof structure $\Pi$ and a switching $s$ we obtain a \emph{correction graph} $G$ by
\begin{itemize}
\item replacing each par link by an edge connecting the conclusion of the link to the premiss selected by $s$
\item replacing each universal link by an edge connecting the conclusion of the link to the formula selected by $s$
\end{itemize}
\end{definition}

Whereas a proof structure is a graph with some additional structure (paired edges, draw as connected dotted lines for the par links, and ``universal'' edges, draw as dotted lines) a correction graph is a plain graph as used in graph theory: both types of special edges are replaced by normal edges according to the switching $s$.

\begin{definition}
A proof structure is a \emph{proof net} iff for all switchings $s$ the corresponding correction graph $G$ is acyclic and connected.
\end{definition}

Remarkably, the proof nets correspond exactly to the provable statements in MILL1 \cite{quant}.

The basic idea of \cite{mill1} is very simple: instead of using the well-known translation of Lambek calculus formulas into first-order logic \citeaffixed{dosen92frames}{used for model-theory, see e.g.}, we use this same translation to obtain formulas of first-order multiplicative \emph{linear} logic. In this paper, I extend this result to the discontinuous Lambek calculus D, while at the same time sketching some novel applications of the system which correspond more closely to analyses in multimodal categorial grammars.

%This means we can get some results directly: since MILL1 is NP complete, we get that (universal) recognition for the Lambek calculus is in NP without having to employ a guarded fragment argument. It also allows us to use the proof-theoretic innovations of linear logic, such as proof nets for the Lambek calculus (something realized before, eg.\ in \cite{Roorda}, but which is now a simple consequence of the translation) 

\subsection{A Danos-style correctness condition}
\label{sec:contract}

Though the correctness condition is conceptually simple, a proof structure has a number of correction graphs which is exponential in the number of asynchronous links, making the correctness condition hard to verify directly (though linear-time algorithms for checking the correctness condition exist in the quantifier-free case, eg. \cite{pnlinear,murong}).

Here, I present an extension of the correctness condition of \cite{reductions} to the first-order case, which avoids this exponential complexity.
Let $G$ be a proof structure, where each vertex of the proof structure is a assigned the set of eigenvariables which occur in the corresponding formula. Then we have the following contractions.

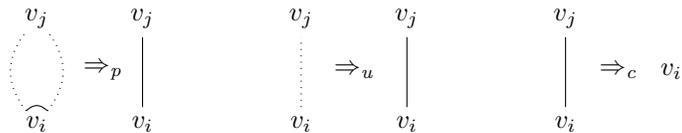
\begin{center}
\begin{tikzpicture}
\node (x) at (0em,0em) {$v_i$};
\node (y) at (0em,4em) {$v_j$};
\draw [dotted] plot [smooth, tension=1] coordinates {(-0.4em,0.5em) (-1em,2em) (-0.4em,3.5em)};
\draw [dotted] plot [smooth, tension=1] coordinates {(0.4em,0.5em) (1em,2em) (0.4em,3.5em)};
\draw plot [smooth,tension=1] coordinates {(-0.4em,0.5em) (0em,0.7em) (0.4em,0.5em)};
\node (x2) at (4em,0em) {$v_i$};
\node (y2) at (4em,4em) {$v_j$};
\draw (x2) -- (y2);
\node (a1) at (2.5em,2em) {$\Rightarrow_{\textit{p}}$};
\node (x3) at (10em,0em) {$v_i$};
\node (y3) at (10em,4em) {$v_j$};
\draw [dotted] (x3) -- (y3);
\node (x4) at (14em,0em) {$v_i$};
\node (y4) at (14em,4em) {$v_j$};
\draw (x4) -- (y4);
\node (a2) at (12em,2em) {$\Rightarrow_{\textit{u}}$};
\node (x3) at (20em,0em) {$v_i$};
\node (y3) at (20em,4em) {$v_j$};
\draw (x3) -- (y3);
\node (x4) at (24em,2em) {$v_i$};
\node (a3) at (22em,2em) {$\Rightarrow_{\textit{c}}$};
\end{tikzpicture}
\end{center}

There is one contraction for the par links ($p$), one contraction for the universal links ($u$) and a final contraction which contracts components (connected subgraphs consisting only of synchronous, axiom and cut links) to a single vertex ($c$).
The $u$ contraction has the condition that there are no occurrences of the eigenvariable of the universal variable corresponding to the link outside of $v_j$. The $c$ contraction has as condition that $i \neq j$; it contracts the vertex connecting $i$ and $j$ and the set of eigenvariables of $v_i$ on the right hand side of the contraction corresponds to the set union of the eigenvariables of $v_i$ and $v_j$ on the left hand side of the contraction.

The following proposition is easy to prove using induction on the number of asynchronous links in the proof structure, using a variant of the ``splitting par'' sequentialization proof of \citeasnoun{reductions}

\begin{proposition}
A proof structure is a proof net iff it contracts to a single vertex using the contractions $p$, $u$ and $c$.
\end{proposition}

It is also easy to verify that the contractions are confluent, and can therefore be applied in any desired order.

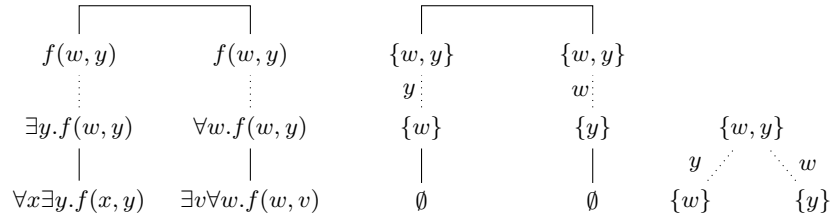
\begin{figure}
\begin{center}
\begin{tikzpicture}
\node (a) at (0em,0em) {$\forall x \exists y. f(x,y)$};
\node (b) at (0em,3em) {$\exists y. f(w,y)$};
\node (c) at (0em,6em) {$f(w,y)$};
\node (e) at (7em,6em) {$f(w,y)$};
\node (f) at (7em,3em) {$\forall w. f(w,y)$};
\node (d) at (7em,0em) {$\exists v \forall w. f(w,v)$};
\draw (d) -- (f);
\draw [dotted] (f) -- (e);
\draw [dotted] (b) -- (c);
\draw (a) -- (b);
\draw (e) -- (7em,8em) -- (0em,8em) -- (c);
\node (a) at (14em,0em) {$\emptyset$};
\node (b) at (14em,3em) {$\{ w \}$};
\node (c) at (14em,6em) {$\{ w, y\}$};
\node (e) at (21em,6em) {$\{ w, y\}$};
\node (f) at (21em,3em) {$\{ y \}$};
\node (d) at (21em,0em) {$\emptyset$};
\draw (d) -- (f);
\draw [dotted] (f) -- (e);
\draw [dotted] (b) -- (c);
\draw (a) -- (b);
\draw (e) -- (21em,8em) -- (14em,8em) -- (c);
\node (y) at (13.5em,4.5em) {$y$};
\node (w) at (20.5em,4.5em) {$w$};
\node (g) at (27.5em,3em) {$\{ w,y \}$};
\node (h) at (25em,0em) {$\{ w \}$};
\node (i) at (30em,0em) {$\{ y \}$};
\draw [dotted] (h) -- (g);
\draw [dotted] (i) -- (g);
\node (y) at (25.2em,1.5em) {$y$};
\node (w) at (29.8em,1.5em) {$w$};
\end{tikzpicture}
\end{center}
\caption{Proof structure and partial contraction sequence for the underivable statement $\forall x \exists y. f(x,y) \vdash \exists v \forall w. f(w,v)$}
\label{fig:exa}
\end{figure}

To give an idea of how these contractions are applied, Figure~\ref{fig:exa} shows (on the left) a proof structure for the underivable statement $\forall x \exists y. f(x,y) \vdash \exists v \forall w. f(w,v)$. In the middle of the Figure, we see the proof structure with each formula replaced by the set of its free variables and before any contractions, with the eigenvariables shown next to their universal links. On the right, we see the structure after all $c$ contractions have been applied. It is clear that we cannot apply the $u$ contraction for $y$, since $y$ occurs at a vertex other than the top vertex. Similarly, we cannot apply the $u$ contraction for $w$ either, meaning the proof structure is not contractible and therefore not a proof net.

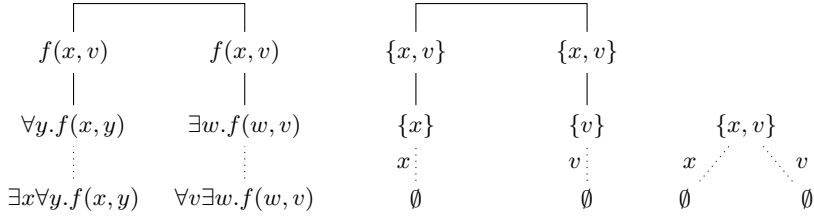
\begin{figure}
\begin{center}
\begin{tikzpicture}
\node (a) at (0em,0em) {$\exists x \forall y. f(x,y)$};
\node (b) at (0em,3em) {$\forall y. f(x,y)$};
\node (c) at (0em,6em) {$f(x,v)$};
\node (e) at (7em,6em) {$f(x,v)$};
\node (f) at (7em,3em) {$\exists w. f(w,v)$};
\node (d) at (7em,0em) {$\forall v \exists w. f(w,v)$};
\draw [dotted] (d) -- (f);
\draw (f) -- (e);
\draw (b) -- (c);
\draw [dotted] (a) -- (b);
\draw (e) -- (7em,8em) -- (0em,8em) -- (c);
\node (a) at (14em,0em) {$\emptyset$};
\node (b) at (14em,3em) {$\{ x \}$};
\node (c) at (14em,6em) {$\{ x,v \}$};
\node (e) at (21em,6em) {$\{ x,v \}$};
\node (f) at (21em,3em) {$\{ v \}$};
\node (d) at (21em,0em) {$\emptyset$};
\node (x) at (13.5em,1.5em) {$x$};
\node (v) at (20.5em,1.5em) {$v$};
\draw [dotted] (d) -- (f);
\draw (f) -- (e);
\draw (b) -- (c);
\draw [dotted] (a) -- (b);
\draw (e) -- (21em,8em) -- (14em,8em) -- (c);
\node (g) at (27.5em,3em) {$\{ x,v \}$};
\node (h) at (25em,0em) {$\emptyset$};
\node (i) at (30em,0em) {$\emptyset$};
\draw [dotted] (h) -- (g);
\draw [dotted] (i) -- (g);
\node (y) at (25.2em,1.5em) {$x$};
\node (w) at (29.8em,1.5em) {$v$};
\end{tikzpicture}
\end{center}
\caption{Proof net and partial contraction sequence for the derivable statement $\exists x \forall y. f(x,y) \vdash \forall v \exists w. f(w,v)$}
\label{fig:exb}
\end{figure}

Figure~\ref{fig:exb} shows the proof structure and part of the contraction sequence for the derivable statement $\exists x \forall y. f(x,y) \vdash \forall v \exists w. f(w,v)$. In this case, the structure on the right \emph{does} allow us to perform the $u$ contractions (in any order), producing a single vertex and thereby showing the proof structure is a proof net.

\subsection{Eager Application of the Contractions}
\label{sec:eager}

Though the contraction condition can be efficiently implemented, when verifying whether or not a given statement is a proof net it is often possible to disqualify partial proof structures (that is, proof structures where only some of the axiom links have been performed). Since the number of potential axiom links is enormous ($n!$ in the worst case), efficient methods for limiting the combinatorial explosion as much as possible are a prerequisite for performing proof search on realistic examples.

The contractions allow us to give a compact representation of the search space by reducing the partial proof structure produced so far. When each vertex is assigned a multiset of literals (in addition to the set of eigenvariables already required for the contractions), the axiom rule corresponds to selecting, if necessary, while unifying the existentially quantified variables, two conjugate literals $+A$ and $-A$ from two different vertices (since the axiom rule corresponds to an application of the $c$ contraction), identifying the two vertices and taking the multiset union of the remaining literals from the two vertices, in addition to taking the set union of the eigenvariables of the vertices. When the input consists of (curried) Horn clauses, each vertex will correspond to a Horn clause; therefore this partial proof structure approach generalizes resolution theorem proving. However, it allows for a lot of freedom in the strategy of literal selection, so we can apply ``smart backtracking'' strategies such as selecting the literal which has the smallest number of conjugates \cite{moot07filter}. The contraction condition immediately suggest the following.

\begin{itemize}
\item never connect a literal to a descendant or an ancestor (generalizes ``formulas from different vertices'' for the Horn clause case); failure to respect this constraint will result in a cyclic proof structure,
\item if the premiss of an asynchronous link is a leaf with the empty set of literals, then we must be able to contract it immediately ; failure to respect this constraint will result in a disconnected proof structure.
\item similarly, if an isolated vertex which is not the only vertex in the graph has the empty set of literals, then the proof structure is disconnected.
\end{itemize}

\section{The Displacement calculus}\label{sec:d}

The Displacement calculus \cite{mv11displacement} is an extension of the Lambek calculus using tuples of strings as their basic units.

\subsection{String tuples}

Whereas the Lambek calculus is the logic of strings, several
formalisms are using \emph{tuples} of strings as their basic units (eg.\ MCFGs, RCGs).

In what follows I use $s$, $s_0,s_1,\ldots, s', s'', \ldots$ to refer
to \emph{simple} strings (ie.\ the 1-tuples) with the constant
$\epsilon$ for the empty string. The letters $t,u,v$ etc. refer to
$i$-tuples of strings for $i \geq 1$. I will write a $i$-tuple of strings
as $s_1,\ldots,s_i$, but also (if $i\geq 2$) as $s_1,t$ or $t',s_i$ where $t$ is
understood to be the string tuple $s_2,\ldots,s_i$ and $t'$ the string
tuple $s_1,\ldots,s_{i-1}$, both $(i-1)$-tuples.

The basic operation for simple strings is concatenation. How does this
operation extend to string tuples? For our current purposes, the
natural extension of concatenation to string tuples is the following

$$
(s_1,\ldots,s_m) \,\circ\, (s'_1,\ldots,s'_n) = s_1,\ldots,s_ms'_1,\ldots,s'_n
$$

\noindent where $s_ms'_1$ is the string concatenation of the two
simple strings $s_m$ and $s'_1$. In other words, the result of
concatenating an $m$-tuple $t$ and an $n$-tuple $u$ is the $n+m-1$
tuple obtained by first taking the first $m-1$ elements of $t$, then
the simple string concatenation of the last element of $t$ with the
first element of $u$ and finally the last $n-1$ elements of $u$. When
both $t$ and $u$ are simple strings, then their concatenation is the
string concatenation of their single element.\footnote{Another natural
  way to define concatenation is as point-wise concatenation of the
  different elements of two (equal-sized) tuples, as done by \citeasnoun{stabler03tupled}.}
In what follows, I will simply write $tu$ for the
concatenation of two string tuples $t$ and $u$ and $u[t]$ to
abbreviate $u_1tu_2$.

\subsection{Position pairs}
\label{sec:pospairs}

As is common in computational linguistics, it is sometimes more
convenient to represent a simple string as a pair of string positions,
the first element of the pair representing the leftmost string
position and the second element its rightmost position. These
positions are commonly represented as integers (to make the implicit
linear precedence relation more easily visible). Likewise, we can
represent an $n$-tuple of strings as a $2n$ tuple of string positions.
This representation has the advantage that it makes string
concatenation trivial: if $x_0,x_1$ is a string starting at position $x_0$ and
ending at position $x_1$ and $x_1,x_2$ is a string starting at
position $x_1$ and ending at position $x_2$ then the concatenation of
these two strings is simply $x_0,x_2$ (this is the familiar difference list concatenation from Prolog \cite{PS87}). 

%Defining concatenation this way presupposes that we already know that
%the two strings are adjacent. When parsing, we already know the
%input string $w_1,\ldots,w_n$ and we can assign the words in the input
%string the positions $n_{i-1},n_i$, as shown below (we will treat this
%example sentence in more detail later).

\begin{definition} We say a grammar is \emph{simple in the input string} if for each input string $w_1,\ldots,w_n$ we have that $w_i$ spans positions ${i},{i+1}$
\end{definition}

Much of the work in parsing presupposes grammars are simple in the input string \cite{ns10parsing}, since it makes the definition of the standard parsing algorithms much neater. However, the original construction of \citeasnoun{cfgfsa} on which it is based is much more general: it computes the intersection of a context-free grammar and a finite-state automaton (FSA), where each non-terminal is assigned an input state and an output state of the FSA. For grammars which are not simple in the input string, this FSA can have self-loops and complex cycles, whereas the input string for a simple grammar is an FSA with a simple, deterministic linear path as shown in the example below. With the exception of Section~\ref{sec:synth}, where I discusses the possibility of abandoning this constraint, the grammars I use will be simple in the input string. A simple example is shown below.

{\samepage 
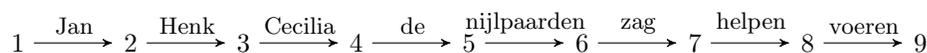
\label{fig:nijlp}
\begin{tikzpicture}[->,>=stealth',shorten >=1pt,auto,node distance=1.5cm]
%\node (a) {0};
\node (b) {1};
\node (c) [right of=b] {2};
\node (d) [right of=c] {3};
\node (e) [right of=d] {4};
\node (f) [right of=e] {5};
\node (g) [right of=f] {6};
\node (h) [right of=g] {7};
\node (i) [right of=h] {8};
\node (j) [right of=i] {9};
%\path (a) edge node {\small dat} (b);
\path (b) edge node {\small Jan} (c);
\path (c) edge node {\small Henk} (d);
\path (d) edge node {\small Cecilia} (e);
\path (e) edge node {\small de} (f);
\path (f) edge node {\small nijlpaarden} (g);
\path (g) edge node {\small zag} (h);
\path (h) edge node {\small helpen} (i);
\path (i) edge node {\small voeren} (j);
\end{tikzpicture}}

Suppose ``nijlpaarden'' (hippos) above is assigned the category
$n$, for noun. Incorporating its string positions produces
$n(5,6)$. It gets more interesting with the determiner ``de'' (the):
we assign it the formula $\forall x. n(5,x) \multimap np(4,x)$, which
says that whenever it finds an $n$ to its immediate right (starting at
position 5 an ending at any
posistion $x$ it will return an $np$ from position 4 to this same
$x$ (this is the MILL1 translation of $np/n$ at position $4,5$). In a chart parser, we would indicate this by adding an $np$ arc
from 4 to 6. There is an important difference with a standard chart parser though: since we are operating in a resource-conscious logic, we know that in a correct proof each rule is used exactly once (though their order is only partially determined).

\begin{figure}
\begin{tikzpicture}
\node at (-9em,2.25em) {Concatentation of $t$ and $u$};
\node at (-8.5em,0.75em) {$t$ an $i$-tuple and $u$ a $j$-tuple};
\draw  (2.5em,1em) rectangle (5em,2em) ;
\draw  (7em,1em) rectangle (9.5em,2em) ;
\draw [fill=gray!30] (9.5em,1em) rectangle (12em,2em) ;
\draw [fill=gray!30] (14em,1em) rectangle (16.5em,2em) ;
\node at (6.0em,1.5em) {$\cdots$};
\node at (13.0em,1.5em) {$\cdots$};
\node at (6.0em,0.0em) {$\underbrace{\qquad\qquad\qquad\ \ \,}_{t}$};
\node at (13.0em,0.0em) {$\underbrace{\qquad\qquad\qquad\ \ \,}_{u}$};
\node at (9.5em,3.6em) {$\overbrace{\qquad\qquad\qquad\qquad\qquad\qquad\qquad}^{tu}$};
\node at (2.5em,2.5em) {$\scriptstyle x_0$};
\node at (9.5em,2.5em) {$\scriptstyle x_n$};
\node at (17.25em,2.5em) {$\scriptstyle x_{n+m}$};
\node at (9em,-2em) {$\scriptstyle n = 2i-1, m=2j-1$};
\end{tikzpicture}

\begin{tikzpicture}
\node at (-10em,2.25em) {Left wrap of $s,t$ and $u$};
\node at (-8.75em,0.75em) {$t$ an $i$-tuple and $u$ a $j$-tuple};
\draw (0em,1em) rectangle (2.5em,2em) ;
\draw [fill=gray!30] (2.5em,1em) rectangle (5em,2em) ;
\draw [fill=gray!30] (7em,1em) rectangle (9.5em,2em) ;
\draw (9.5em,1em) rectangle (12em,2em) ;
\draw (14em,1em) rectangle (16.5em,2em) ;
\node at (6.0em,1.5em) {$\cdots$};
\node at (13.0em,1.5em) {$\cdots$};
\node at (1.25em,0.0em) {$\underbrace{\qquad\ }_{s}$};
\node at (13.0em,0.0em) {$\underbrace{\qquad\qquad\qquad\ \ \,}_{t}$};
\node at (6.0em,0.0em) {$\underbrace{\qquad\qquad\qquad\ \ \,}_{u}$};
\node at (8.25em,3.6em) {$\overbrace{\qquad\qquad\qquad\qquad\qquad\qquad\qquad\qquad\ }^{sut}$};
\node at (0em,2.5em) {$\scriptstyle x_0$};
\node at (2.5em,2.5em) {$\scriptstyle x_1$};
\node at (9.5em,2.5em) {$\scriptstyle x_n$};
\node at (17.25em,2.5em) {$\scriptstyle x_{n+m}$};
\node at (9em,-2em) {$\scriptstyle n = 2j, m=2i-1$};
\end{tikzpicture}

\begin{tikzpicture}
\node at (-9.75em,2.25em) {Right wrap of $t,s$ and $u$};
\node at (-8.75em,0.75em) {$t$ an $i$-tuple and $u$ a $j$-tuple};
\draw (16.5em,1em) rectangle (14.0em,2em) ;
\draw [fill=gray!30] (14.0em,1em) rectangle (11.5em,2em) ;
\draw [fill=gray!30] (9.5em,1em) rectangle (7em,2em) ;
\draw (7em,1em) rectangle (4.5em,2em) ;
\draw (2.5em,1em) rectangle (0em,2em) ;
\node at (10.5em,1.5em) {$\cdots$};
\node at (3.5em,1.5em) {$\cdots$};
\node at (15.25em,0.0em) {$\underbrace{\qquad\ }_{s}$};
\node at (3.5em,0.0em) {$\underbrace{\qquad\qquad\qquad\ \ \,}_{t}$};
\node at (10.5em,0.0em) {$\underbrace{\qquad\qquad\qquad\ \ \,}_{u}$};
\rmindicess
\node at (9em,-2em) {$\scriptstyle n=2i-1, m = 2j$};
\end{tikzpicture}
\caption{String operations and their equivalent string positions operations}
\label{fig:stringops}
\end{figure}
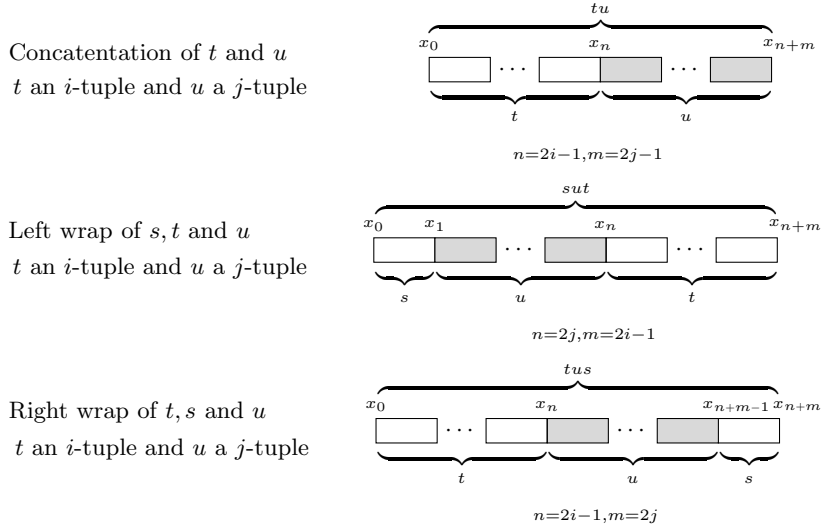

Figure~\ref{fig:stringops} shows the three elementary string
operations of the Displacement calculus both in the form of operations of string
tuples and in the form of operations of string position pairs.

\emph{Concatenation} takes an $i$-tuple $t$ (shown in the top of the figure as
the white blocks, with corresponding string
positions $x_0,\ldots,x_n$ for $n=2i-1$) and a $j$-tuple $u$ (shown in
the top of the figure as the gray blocks, with
corresponding string positions $x_n,\ldots,x_{n+m}$ for $m=2j-1$) and
the resulting concatenation $tu$ (with the last element of $t$
concatenated to the first element of $u$, indicated as the gray-white
block $x_{n-1},x_{n+1}$; $x_n$ is not a string position in the
  resulting $i+j-1$-tuple $tu$, which consists of the string positions $x_0,\ldots,x_{n-1},x_{n+1},\ldots,x_{n+m}$.

\emph{Left wrap} takes an $i+1$-tuple $s,t$ (with $s$ a simple string and
$t$ an $i$-tuple, with string positions $x_0,x_1,x_n,\ldots,x_{n+m}$)
and a $j$-tuple $u$ (with string positions $x_1,\ldots,x_n$) and wraps $s,t$ around $u$
producing and $i+j-1$-tuple $sut$ with string positions
$x_0,x_2,\ldots,x_{n-1},x_{n+1},\ldots,x_{n+m}$, with positions $x_1$
and $x_n$ removed because of the two concatenations.

Symmetrically, \emph{right wrap} takes an $i+1$-tuple $t,s$ (with $s$ a
simple string and $t$ an $i$-tuple) and a $j$-tuple $u$ and wraps
$t,s$ around $u$ producing $tus$.

Given these operations, the proof rules for D are simple to state. I
give a notational variant of the natural deduction calculus of \cite{mv11displacement}.
As usual, natural deduction proofs start with a hypothesis
$t:A$ (for $t:A$ a member of the lexicon of the grammar, in which case
$t$ is a constant from the lexicon, or for a hypothesis discharged by the product
elimination and implication introduction rules). In each case the
constant string $t$ is unique in the proof. 
%Given an input string $s$,
%the lexical hypotheses must span $s$; that is, $s$ is exactly the
%concatenation of all string components of all lexical hypotheses.

 For a given
(sub-)proof, the \emph{active} hypotheses are all hypotheses which
have not been discharged by a product elimination of implication
introduction rule in this (sub-)proof.

For the logical rules, we can see that the different families of
connectives correspond to the three basic string tuple operations:
with concatenation for $/$, $\bullet$ and $\backslash$ (the rules are shown in Figure~\ref{fig:ndlambek}
 and Figure~\ref{fig:concat} with the corresponding string tuples), left wrap for
$\upl$, $\prodl$ and $\downl$ (shown in Figure~\ref{fig:ndlw} and Figure~\ref{fig:lwrap}) and right wrap for $\upr$, $\prodr$
and $\downr$ (shown in Figure~\ref{fig:ndrw}  and Figure~\ref{fig:rwrap}).

\begin{figure}

$$
\begin{array}{ccc}
\infer[\backslash E]{tu:C}{t:A & u:A\mathbin{\backslash} C} &&
\infer[\backslash I_i]{u:A\mathbin{\backslash} C}{\infer*{tu:C}{[ t:A ]^i}}\\
\\
\infer[/ E]{tu:C}{t:C\mathbin{/} B & u:B} &&
\infer[/ I_i]{t:C\mathbin{/} B}{\infer*{tu:C}{[ u:B ]^i}}
\\
\\
\infer[\bullet E_i]{u[t]:C}{t:A\bullet B & \infer*{u[t_1t_2]:C}{[t_1:A]^i & [t_2:B]^i}}
&&
\infer[\bullet I]{tu:A\bullet B}{t:A & u:B}
\end{array}
$$

\caption{Proof rules -- Lambek calculus}
\label{fig:ndlambek}
\end{figure}

\begin{figure}

$$
\begin{array}{ccc}
\infer[\downl E]{sut:C}{s,t:A & u:A\downl C} &&
\infer[\downl I_i]{u:A\downl C}{\infer*{sut:C}{[s,t:A]^i}}
\\
\\
\infer[\upl E]{sut:C}{s,t:C\upl B & u:B} &&
\infer[\upl I_i]{s,t:C\upl B}{\infer*{sut:C}{[u:B]^i}}
\\
\\
\infer[\prodl E_i]{u[t]:C}{t:A\prodl B & \infer*{u[st_1t_2]:C}{[s,t_2:A]^i & [t_1:B]^i}}
&&
\infer[\prodl I]{sut:A\prodl B}{s,t:A & u:B}
\end{array}
$$

\caption{Proof rules --- leftmost infixation,extraction}
\label{fig:ndlw}
\end{figure}

\begin{figure}

$$
\begin{array}{ccc}
\infer[\downr E]{sut:C}{t,s:A & u:A\downr C} &&
\infer[\downr I_i]{u:A\downr C}{\infer*{tus:C}{[t,s:A]^i}}\\
\\
\infer[\upr E]{sut:C}{t,s:C\upr B & u:B} &&
\infer[\upr I_i]{t,s:C\upr B}{\infer*{tus:C}{[u:B]^i}}
\\
\\
\infer[\prodr E_i]{u[t]:C}{t:A\prodr B & \infer*{u[t_1t_2s]:C}{[t_1,s:A]^i & [t_2:B]^i}}
&&
\infer[\prodr I]{tus:A\prodr B}{t,s:A & u:B}
\end{array}
$$

\caption{Proof rules --- rightmost infixation,extraction}
\label{fig:ndrw}
\end{figure}

In the discontinuous Lambek calculus, we define the \emph{sort} of a
formula $F$, written $s(F)$ as the number of items in its string tuple
minus 1. Given sorts for the atomic formulas, we compute the sort of a
complex formula as shown in Table~\ref{tab:sorts} (the distinction
between the left wrap and right wrap connectives is irrelevant for the
sorts).

\begin{table}
\begin{align*}
s(A\mathbin{\bullet} B) &= s(A) + s(B) \\
s(A\mathbin{\backslash} C) &= s(C) - s(A) \\
s(C\mathbin{/}B) &= s(C) - s(B) \\[2mm]
%\end{align*}
%
%\begin{align*}
s(A\mathbin{\odot} B) &= s(A) + s(B) -1 \\
s(A\mathbin{\downarrow} C) &= s(C) + 1 - s(A) \\
s(C\mathbin{\uparrow}B) &= s(C) + 1 - s(B) \\
\end{align*}
\caption{Computing the sort of a complex formula given the sort of its
  immediate subformulas}
\label{tab:sorts}
\end{table}

\subsection{MILL1 and Multiple Context-Free Grammars}
\label{sec:mcfg}

It is fairly easy to see that MILL1 generates (at least) the multiple context-free languages (or equivalently, the languages generated by simple, positive range concatenation grammars \cite{boullier}) by using a lexicalized form of the grammars as defined below. 

\begin{definition}
A grammar is \emph{lexicalized} if each grammar rule uses exactly one non-terminal symbol.
\end{definition}

Lexicalization is one of the principal differences between traditional phrase structure grammars and categorial grammars: categorial grammars generally require a form of lexicalization, whereas phrase structure grammars do not.
The most well-known lexicalized form is the Greibach normal form for context-free grammars \cite{Greibach:1965:NNF}. \citeasnoun{wijnholds} shows that any ($\epsilon$-free) simple, positive range concatenation grammar has a lexicalized grammar generating the same language. Since ranges are simply pairs of non-negative integers (see Section~\ref{sec:pospairs} and \cite{boullier}) these translate directly to Horn clauses in MILL1. The following rules are therefore both a notational variant of a (lexicalized) MCFG/sRCG and an MILL1 lexicon (corresponding to the verbs of the example on page~\pageref{fig:nijlp}). See Appendix~\ref{app:nijl} for the construction of a proof net using (the curried versions of) these formulas.

\vspace{-.5\baselineskip}
\begin{align*}
\forall x_0 x_1 x_2 x_3. np(x_0,x_1) \otimes np(x_1,x_2) \otimes \textit{inf}(x_2,6,7,x_3) &\multimap s(x_0,x_3) & \textit{zag}\\
\forall x_0 x_1 x_2 x_3. np(x_0,x_1) \otimes \textit{inf}(x_1,x_2,8,x_3) &\multimap \textit{inf}(x_0,x_2,7,x_3) & \textit{helpen}\\
\forall x_0 x_1. np(x_0,x_1) &\multimap \textit{inf}(x_0,x_1,8,9) & \textit{voeren}\\
\end{align*}
\vspace{-.5\baselineskip}

In Section~\ref{sec:synth}, we will see how we can obtain these formulas via a translation of D. 

\section{Translations}
\label{sec:trans}
%With all this in place, 

The proof rules and the corresponding string tuple operations (shown in Figures~\ref{fig:concat}, \ref{fig:lwrap} and \ref{fig:rwrap}) suggest
the translation shown in Table~\ref{tab:transl} of D formulas into MILL1 formulas. It is an extension of the translation of Lambek calculus formulas
of \cite{mill1}, while at the same time
extending the translation of \cite{mf08disco} for the simple displacement calculus.

% = Lambek connectives

\begin{figure}
\begin{tabular}{cccc}
\begin{tikzpicture}
\draw  (2.5em,1em) rectangle (5em,2em) ;
\draw  (7em,1em) rectangle (9.5em,2em) ;
\draw [fill=gray!30] (9.5em,1em) rectangle (12em,2em) ;
\draw [fill=gray!30] (14em,1em) rectangle (16.5em,2em) ;
\node at (6.0em,1.5em) {$\cdots$};
\node at (13.0em,1.5em) {$\cdots$};
\node at (6.0em,0.0em) {$\underbrace{\qquad\qquad\qquad\ \ \,}_{A}$};
\node at (13.0em,0.0em) {$\underbrace{\qquad\qquad\qquad\ \ \,}_{B}$};
\node at (9.5em,3.6em) {$\overbrace{\qquad\qquad\qquad\qquad\qquad\qquad\qquad}^{A\mathbin{\bullet}B}$};
\node at (2.5em,2.5em) {$\scriptstyle x_0$};
\node at (9.5em,2.5em) {$\scriptstyle x_n$};
\node at (17.25em,2.5em) {$\scriptstyle x_{n+m}$};
\node at (9em,-2em) {$\scriptstyle n = 2s(A)+1, m=2s(B)+1$};
%\node at (9em,6.5em) {$\ x$};
\end{tikzpicture}
&\raisebox{2.0ex}{\infer[\bullet E_i]{u[t]:C}{t:A\bullet B & \infer*{u[t_1t_2]:C}{[t_1:A]^i & [t_2:B]^i}}}
&&
\raisebox{2.0ex}{\infer[\bullet I]{tu:A\bullet B}{t:A & u:B}}
 \\
\begin{tikzpicture}
\draw  (2.5em,1em) rectangle (5em,2em) ;
\draw  (7em,1em) rectangle (9.5em,2em) ;
\draw [fill=gray!30] (9.5em,1em) rectangle (12em,2em) ;
\draw [fill=gray!30] (14em,1em) rectangle (16.5em,2em) ;
\node at (6.0em,1.5em) {$\cdots$};
\node at (13.0em,1.5em) {$\cdots$};
\node at (6.0em,-0.1em) {$\underbrace{\qquad\qquad\qquad\ \ \,}_{C\mathbin{/}B}$};
\node at (13.0em,0.0em) {$\underbrace{\qquad\qquad\qquad\ \ \,}_{B}$};
\node at (9.5em,3.6em) {$\overbrace{\qquad\qquad\qquad\qquad\qquad\qquad\qquad}^{C}$};
\node at (2.5em,2.5em) {$\scriptstyle x_0$};
\node at (9.5em,2.5em) {$\scriptstyle x_n$};
\node at (17.25em,2.5em) {$\scriptstyle x_{n+m}$};
\node at (9em,-2em) {$\scriptstyle n = 2s(C\mathbin{/}B)+1, m=2s(B)+1$};
\end{tikzpicture}
&\raisebox{2.0ex}{\infer[/ E]{tu:C}{t:C\mathbin{/} B & u:B}} &&
\raisebox{2.0ex}{\infer[/ I_i]{t:C\mathbin{/} B}{\infer*{tu:C}{[u:B]^i}}}
  \\
\begin{tikzpicture}
\draw  [fill=gray!30] (2.5em,1em) rectangle (5em,2em) ;
\draw  [fill=gray!30] (7em,1em) rectangle (9.5em,2em) ;
\draw (9.5em,1em) rectangle (12em,2em) ;
\draw (14em,1em) rectangle (16.5em,2em) ;
\node at (6.0em,1.5em) {$\cdots$};
\node at (13.0em,1.5em) {$\cdots$};
\node at (6.0em,0.0em) {$\underbrace{\qquad\qquad\qquad\ \ \,}_{A}$};
\node at (13.0em,-0.1em) {$\underbrace{\qquad\qquad\qquad\ \ \,}_{A\mathbin{\backslash} C}$};
\node at (9.5em,3.6em) {$\overbrace{\qquad\qquad\qquad\qquad\qquad\qquad\qquad}^{C}$};
\node at (2.5em,2.5em) {$\scriptstyle x_0$};
\node at (9.5em,2.5em) {$\scriptstyle x_n$};
\node at (17.25em,2.5em) {$\scriptstyle x_{n+m}$};
\node at (9em,-2em) {$\scriptstyle n = 2s(A)+1, m=2s(A\mathbin{\backslash}C)+1$};
\end{tikzpicture}&
\raisebox{2.0ex}{\infer[\backslash E]{tu:C}{t:A & u:A\mathbin{\backslash} C}} &&
\raisebox{2.0ex}{\infer[\backslash I_i]{u:A\mathbin{\backslash} C}{\infer*{tu:C}{[t:A]^i}}}
\end{tabular}
\caption{String positions -- Lambek calculus}
\label{fig:concat}
\end{figure}

\begin{figure}
\begin{tabular}{cccc}
\begin{tikzpicture}
\draw (0em,1em) rectangle (2.5em,2em) ;
\draw [fill=gray!30] (2.5em,1em) rectangle (5em,2em) ;
\draw [fill=gray!30] (7em,1em) rectangle (9.5em,2em) ;
\draw (9.5em,1em) rectangle (12em,2em) ;
\draw (14em,1em) rectangle (16.5em,2em) ;
\node at (6.0em,1.5em) {$\cdots$};
\node at (13.0em,1.5em) {$\cdots$};
\node at (1.25em,0.0em) {$\underbrace{\qquad\ }_{A}$};
\node at (13.0em,0.0em) {$\underbrace{\qquad\qquad\qquad\ \ \,}_{A}$};
\node at (6.0em,0.0em) {$\underbrace{\qquad\qquad\qquad\ \ \,}_{B}$};
\node at (8.25em,3.6em) {$\overbrace{\qquad\qquad\qquad\qquad\qquad\qquad\qquad\qquad\ }^{A\mathbin{\prodl} B}$};
\node at (0em,2.5em) {$\scriptstyle x_0$};
\node at (2.5em,2.5em) {$\scriptstyle x_1$};
\node at (9.5em,2.5em) {$\scriptstyle x_n$};
\node at (17.25em,2.5em) {$\scriptstyle x_{n+m}$};
\node at (9em,-2em) {$\scriptstyle n = 2s(B)+2, m=2s(A)-1$};
\end{tikzpicture}
&\raisebox{2.0ex}{\infer[\prodl E_i]{u[t]:C}{t:A\prodl B & \infer*{u[st_1t_2]:C}{[s,t_2:A]^i & [t_1:B]^i}}}
&&
\raisebox{2.0ex}{\infer[\prodl I]{sut:A\prodl B}{s,t:A & u:B}}
 \\
\begin{tikzpicture}
\draw (0em,1em) rectangle (2.5em,2em) ;
\draw [fill=gray!30] (2.5em,1em) rectangle (5em,2em) ;
\draw [fill=gray!30] (7em,1em) rectangle (9.5em,2em) ;
\draw (9.5em,1em) rectangle (12em,2em) ;
\draw (14em,1em) rectangle (16.5em,2em) ;
\node at (6.0em,1.5em) {$\cdots$};
\node at (13.0em,1.5em) {$\cdots$};
\node at (1.25em,-0.1em) {$\underbrace{\qquad\ }_{C\mathbin{\upl} B}$};
\node at (13.0em,-0.1em) {$\underbrace{\qquad\qquad\qquad\ \ \,}_{C\mathbin{\upl} B}$};
\node at (6.0em,0.0em) {$\underbrace{\qquad\qquad\qquad\ \ \,}_{B}$};
\node at (8.25em,3.6em) {$\overbrace{\qquad\qquad\qquad\qquad\qquad\qquad\qquad\qquad\ }^{C}$};
\node at (0em,2.5em) {$\scriptstyle x_0$};
\node at (2.5em,2.5em) {$\scriptstyle x_1$};
\node at (9.5em,2.5em) {$\scriptstyle x_n$};
\node at (17.25em,2.5em) {$\scriptstyle x_{n+m}$};
\node at (9em,-2em) {$\scriptstyle n = 2s(B)+2, m=2s(C\mathbin{\upl}B)-1$};
\end{tikzpicture}
&\raisebox{2.0ex}{\infer[\upl E]{sut:C}{s,t:C\upl B & u:B}} &&
\raisebox{2.0ex}{\infer[\upl I_i]{s,t:C\upl B}{\infer*{sut:C}{[u:B]^i}}}
  \\
\begin{tikzpicture}
\draw [fill=gray!30] (0em,1em) rectangle (2.5em,2em) ;
\draw  (2.5em,1em) rectangle (5em,2em) ;
\draw  (7em,1em) rectangle (9.5em,2em) ;
\draw [fill=gray!30] (9.5em,1em) rectangle (12em,2em) ;
\draw [fill=gray!30] (14em,1em) rectangle (16.5em,2em) ;
\node at (6.0em,1.5em) {$\cdots$};
\node at (13.0em,1.5em) {$\cdots$};
\node at (1.25em,0.0em) {$\underbrace{\qquad\ }_{A}$};
\node at (13.0em,0.0em) {$\underbrace{\qquad\qquad\qquad\ \ \,}_{A}$};
\node at (6.0em,-0.1em) {$\underbrace{\qquad\qquad\qquad\ \ \,}_{A\mathbin{\downl} C}$};
\node at (8.25em,3.6em) {$\overbrace{\qquad\qquad\qquad\qquad\qquad\qquad\qquad\qquad\ }^{C}$};
\node at (0em,2.5em) {$\scriptstyle x_0$};
\node at (2.5em,2.5em) {$\scriptstyle x_1$};
\node at (9.5em,2.5em) {$\scriptstyle x_n$};
\node at (17.25em,2.5em) {$\scriptstyle x_{n+m}$};
\node at (9em,-2em) {$\scriptstyle n = 2s(A\mathbin{\downl}C)+2, m=2s(A)-1$};
\end{tikzpicture}&
\raisebox{2.0ex}{\infer[\downl E]{sut:C}{s,t:A & u:A\downl C}} &&
\raisebox{2.0ex}{\infer[\downl I_i]{u:A\downl C}{\infer*{sut:C}{[s,t:A]^i}}}
\end{tabular}
\caption{String positions -- leftmost infix/extraction}
\label{fig:lwrap}
\end{figure}

\begin{figure}
\begin{tabular}{cccc}
\begin{tikzpicture}
\draw (16.5em,1em) rectangle (14.0em,2em) ;
\draw [fill=gray!30] (14.0em,1em) rectangle (11.5em,2em) ;
\draw [fill=gray!30] (9.5em,1em) rectangle (7em,2em) ;
\draw (7em,1em) rectangle (4.5em,2em) ;
\draw (2.5em,1em) rectangle (0em,2em) ;
\node at (10.5em,1.5em) {$\cdots$};
\node at (3.5em,1.5em) {$\cdots$};
\node at (15.25em,0.0em) {$\underbrace{\qquad\ }_{A}$};
\node at (3.5em,0.0em) {$\underbrace{\qquad\qquad\qquad\ \ \,}_{A}$};
\node at (10.5em,0.0em) {$\underbrace{\qquad\qquad\qquad\ \ \,}_{B}$};
%\node at (8.25em,2.9em) {$\overbrace{\qquad\qquad\qquad\qquad\qquad\qquad\qquad\qquad\ }^{A\mathbin{\odot} B}$};
\rmindicesp
\node at (9em,-2em) {$\scriptstyle n=2s(A)-1, m = 2s(B)+2$};
\end{tikzpicture}
&\raisebox{2.0ex}{\infer[\prodr E_i]{u[t]:C}{t:A\prodr B & \infer*{u[t_1t_2s]:C}{[t_1,s:A]^i & [t_2:B]^i}}}
&&
\raisebox{2.0ex}{\infer[\prodr I]{tus:A\prodr B}{t,s:A & u:B}}
 \\
\begin{tikzpicture}
\draw (16.5em,1em) rectangle (14.0em,2em) ;
\draw [fill=gray!30] (14.0em,1em) rectangle (11.5em,2em) ;
\draw [fill=gray!30] (9.5em,1em) rectangle (7em,2em) ;
\draw (7em,1em) rectangle (4.5em,2em) ;
\draw (2.5em,1em) rectangle (0em,2em) ;
\node at (10.5em,1.5em) {$\cdots$};
\node at (3.5em,1.5em) {$\cdots$};
\node at (15.25em,-0.1em) {$\underbrace{\qquad\ }_{C\mathbin{\upr} B}$};
\node at (3.5em,-0.1em) {$\underbrace{\qquad\qquad\qquad\ \ \,}_{C\mathbin{\upr} B}$};
\node at (10.5em,0.0em) {$\underbrace{\qquad\qquad\qquad\ \ \,}_{B}$};
%\node at (8.25em,3.0em)
%{$\overbrace{\qquad\qquad\qquad\qquad\qquad\qquad\qquad\qquad\
%}^{C}$};
\rmindices
\node at (9em,-2em) {$\scriptstyle n=2s(C\mathbin{\upr}B)-1, m = 2s(B)+2$};
\end{tikzpicture}
&\raisebox{2.0ex}{\infer[\upr E]{sut:C}{t,s:C\upr B & u:B}} &&
\raisebox{2.0ex}{\infer[\upr I_i]{t,s:C\upr B}{\infer*{tus:C}{[u:B]^i}}}
\\
\begin{tikzpicture}
\draw [fill=gray!30] (16.5em,1em) rectangle (14.0em,2em) ;
\draw  (14em,1em) rectangle (11.5em,2em) ;
\draw  (9.5em,1em) rectangle (7em,2em) ;
\draw [fill=gray!30] (7em,1em) rectangle (4.5em,2em) ;
\draw [fill=gray!30] (2.5em,1em) rectangle (0em,2em) ;
\node at (10.5em,1.5em) {$\cdots$};
\node at (3.5em,1.5em) {$\cdots$};
\node at (15.25em,0.0em) {$\underbrace{\qquad\ }_{A}$};
\node at (3.5em,0.0em) {$\underbrace{\qquad\qquad\qquad\ \ \,}_{A}$};
\node at (10.5em,-0.1em) {$\underbrace{\qquad\qquad\qquad\ \
    \,}_{A\mathbin{\downr} C}$};
\rmindices
\node at (9em,-2em) {$\scriptstyle n=2s(A)-1, m = 2s(A\mathbin{\downr}C)+2$};
\end{tikzpicture}
&
\raisebox{2.0ex}{\infer[\downr E]{sut:C}{t,s:A & u:A\downr C}} &&
\raisebox{2.0ex}{\infer[\downr I_i]{u:A\downr C}{\infer*{tus:C}{[t,s:A]^i}}}
\end{tabular}
\caption{String positions -- rightmost infix/extraction}
\label{fig:rwrap}
\end{figure}

%\pagebreak
%\subsection{Wrapping and infixation}

The reader intimidated by the number variable indices in Table~\ref{tab:transl} is invited to look at Figures~\ref{fig:concat}, \ref{fig:lwrap} and \ref{fig:rwrap} for the correspondence between the string position numbers and the strings components of the different formulas in the translation. Section~\ref{sec:examples} will illustrate the translation using some examples, whereas Section~\ref{sec:correct} will make the correspondence the rules from the Displacement calculus more precise.

%Table~\ref{tab:transl} lists the translation of D formulas into MILL1 formulas.
Note: the sequence $x_i, \ldots, x_i$ is of course simply the unit sequence $x_i$ whereas
the sequence $x_i,\ldots,x_{i-1}$ is the empty sequence.

\editout{
$$
\| A \prodl B \|^{x_0,x_2,\ldots,x_{n-1},x_{n+1},\ldots,x_{n+m}} =
  \exists x_1, x_n \| A \|^{x_0,x_1,x_n,\ldots,x_{n+m}} \otimes \| B \|^{x_1,\ldots,x_n}
$$

Simple discontinuity: $n=2$, $m=1$

$$
\| A \prodr B \|^{x_0,\ldots,x_{n-1},x_{n+1},\ldots,x_{n+m-2},x_{n+m}} =
  \exists x_n, x_{n+m-1} \| A \|^{x_0,\ldots,x_n,x_{n+m-1},x_{n+m}} \otimes \| B \|^{x_n,\ldots,x_{n+m-1}}
$$

Simple discontinuity: $n=1$, $m=2$

$$
\| A \odot B \|^{x_0,x_{3}} =
  \exists x_1, x_2 \| A \|^{x_0,x_1,x_2,x_{3}} \otimes \| B \|^{x_1,x_2}
$$

$$
\| C \mathbin{\upl} B \|^{x_0,x_1,x_n, \ldots, x_{n+m}} = \forall
x_2 ,\ldots x_{n-1} \| B \|^{x_1,\ldots,x_{n}} \multimap \|C \|^{x_0,x_2\ldots,x_{n-1},x_{n+1},\ldots,x_{n+m}}
$$

%Simple discontinuity: $n=2$, $m=1$

$$
\| C \mathbin{\upr} B \|^{x_0,\ldots,x_n, x_{n+m-1}, x_{n+m}} = \forall
x_{n+1} ,\ldots x_{n+m-2} \| B \|^{x_n,\ldots,x_{n+m-1}} \multimap \|C \|^{x_0,\ldots,x_{n-1},x_{n+1},\ldots,x_{n+m-2},x_{n+m}}
$$

%Simple discontinuity: $n=1$, $m=2$

$$
\| A \mathbin{\downl} C \|^{x_1, \ldots, x_{n}} = \forall x_0 ,x_{n+1}
\ldots, x_{n+m} \| A \|^{x_0,x_1,x_n,\ldots,x_{n+m}} \multimap \|C \|^{x_0,x_2\ldots,x_{n-1},x_{n+1},\ldots,x_{n+m}}
$$

%Simple discontinuity: $n=2$, $m=1$

$$
\| A \mathbin{\downr} C \|^{x_1, \ldots, x_{n}} = \forall x_0 ,\ldots,x_{n-1}, x_{n+m} \| A \|^{x_0,\ldots,x_n,x_{n+m-1},x_{n+m}} \multimap \|C \|^{x_0,\ldots,x_{n-1},x_{n+1},\ldots,x_{n+m-2},,x_{n+m}}
$$

%Simple discontinuity: $n=1$, $m=2$. 
}

\begin{table}
\begin{flalign}\label{eq:prod}
\left.\begin{aligned}[l]
&  \| A \mathbin{\bullet} B\|^{x_0,\ldots,x_{n-1},x_{n+1},\ldots,x_{n+m}}  \\
& \quad = \exists x_n \| A \|^{x_0,\ldots,x_n} \otimes \| B
\|^{x_n,\ldots,x_{n+m}}\end{aligned} \right\rbrace \scriptstyle
n=2s(A)+1, m=2s(B)+1\ \ \,\,\, \\
\label{eq:dr}\left.\begin{aligned}[l]
&\| C \mathbin{/} B \|^{x_0,\ldots,x_n}  \\
&\quad = \forall x_{n+1},\ldots,
x_{n+m} \| B \|^{x_n,\ldots,x_{n+m}} \multimap\\ & \qquad\qquad\qquad\qquad\qquad \| C
\|^{x_0,\ldots,x_{n-1},x_{n+1},\ldots, x_{n+m}}
\end{aligned} \right\rbrace \scriptstyle  n=2s(C\mathbin{/} B)+1, m=2s(B)+1 \\
\label{eq:dl}\left.\begin{aligned}[l]
&\| A \mathbin{\backslash} C \|^{x_n, \ldots, x_{n+m}}  \\
&\quad = \forall x_0 ,
\ldots, x_{n-1} \| A \|^{x_0,\ldots,x_n} \multimap\\ & \qquad\qquad\qquad\qquad\qquad \|C
\|^{x_0,\ldots,x_{n-1},x_{n+1},\ldots,x_{n+m}}
\end{aligned} \right\rbrace \scriptstyle  n=2s(A)+1, m=2s(A\mathbin{\backslash}C)+1 
\end{flalign}

\begin{flalign}\label{eq:prodl}
\left.\begin{aligned}[l]
& \| A \prodl B \|^{x_0,x_2,\ldots,x_{n-1},x_{n+1},\ldots,x_{n+m}} \\
& \quad = \exists x_1, x_n \| A \|^{x_0,x_1,x_n,\ldots,x_{n+m}} \otimes \| B \|^{x_1,\ldots,x_n}
\end{aligned} \right\rbrace \scriptstyle
n=2s(B)+2, m=2s(A)-1\ \ \ \;\;\; \\
\label{eq:upl}\left.\begin{aligned}[l]
&\| C \mathbin{\upl} B \|^{x_0,x_1,x_n, \ldots, x_{n+m}}  \\
&\quad = \forall
x_2 ,\ldots, x_{n-1} \| B \|^{x_1,\ldots,x_{n}} \multimap\\ &
\qquad\qquad\qquad\qquad\qquad\qquad \|C \|^{x_0,x_2,\ldots,x_{n-1},x_{n+1},\ldots,x_{n+m}}
\end{aligned} \right\rbrace \scriptstyle  n=2s(B)+2, m=2s(C \mathbin{\upl} B)-1 \\%\begin{split}
\label{eq:downl}\left.\begin{aligned}[l]
&\| A \mathbin{\downl} C \|^{x_1, \ldots, x_{n}}  \\
&\quad = \forall x_0 ,x_{n+1},
\ldots, x_{n+m} \| A \|^{x_0,x_1,x_n,\ldots,x_{n+m}} \multimap\\ &
\qquad\qquad\qquad\qquad\qquad\qquad \|C \|^{x_0,x_2,\ldots,x_{n-1},x_{n+1},\ldots,x_{n+m}}
\end{aligned} \right\rbrace \scriptstyle  n=2s(A \mathbin{\downl} C)+2, m=2s(A)-1 
\end{flalign}

\begin{flalign}\label{eq:prodr}
\left.\begin{aligned}[l]
& \| A \prodr B \|^{x_0,\ldots,x_{n-1},x_{n+1},\ldots,x_{n+m-2},x_{n+m}} \\
& \quad = \exists x_n, x_{n+m-1} \| A \|^{x_0,\ldots,x_n,x_{n+m-1},x_{n+m}} \otimes \| B \|^{x_n,\ldots,x_{n+m-1}} \end{aligned} \right\rbrace \scriptstyle
n=2s(A)-1, m=2s(B)+2\ \ \ \ \;\; \\
\label{eq:upr}\left.\begin{aligned}[l]
&\| C \mathbin{\upr} B \|^{x_0,\ldots,x_n, x_{n+m-1}, x_{n+m}}  \\
& \quad = \forall
x_{n+1} ,\ldots, x_{n+m-2} \| B \|^{x_n,\ldots,x_{n+m-1}} \multimap\\ &
\qquad\qquad\qquad\qquad\qquad\qquad \|C \|^{x_0,\ldots,x_{n-1},x_{n+1},\ldots,x_{n+m-2},x_{n+m}}
\end{aligned} \right\rbrace \scriptstyle  n=2s(C \mathbin{\upl} B)-1, m=2s(B)+2 \\%\begin{split}
\label{eq:downr}\left.\begin{aligned}[l]
&\| A \mathbin{\downr} C \|^{x_n, \ldots, x_{n+m-1}}  \\
&\quad = \forall x_0 ,\ldots,x_{n-1}, x_{n+m} \| A \|^{x_0,\ldots,x_n,x_{n+m-1},x_{n+m}} \multimap\\ &
\qquad\qquad\qquad\qquad\qquad\qquad \|C \|^{x_0,\ldots,x_{n-1},x_{n+1},\ldots,x_{n+m-2},x_{n+m}}
\end{aligned} \right\rbrace \scriptstyle  n=2s(A)-1, m= 2s(A \mathbin{\downl} C)+2
\end{flalign}
\caption{Translation of D formulas to MILL1 formulas}
\label{tab:transl}
\end{table}

If there are at most two string tuples, both $ C \mathbin{\upl} B$ (Equation~\ref{eq:upl} with $n=2$, $m=1$, remembering
that $x_2,\ldots,x_{n-1} \equiv x_2,\ldots,x_1$ which is equivalent to
the empty sequence of string positions and the empty sequence of
quantifier prefixes, and that
$x_{n+1},\ldots,x_{n+m} \,\equiv\, x_3,\ldots,x_3 \,\equiv\, x_3$) and $
 C \mathbin{\upr} B$ (Equation~\ref{eq:upr} with $n=1$, $m=2$)  translate to the following

$$
\| C \mathbin{\uparrow} B \|^{x_0,x_1,x_2, x_3} =  \| B \|^{x_1,x_{2}} \multimap \|C \|^{x_0,x_3}
$$

%Where $n$ is $2s(A)+1$ and $m$ is $2s(B)+1$.
%
%$$
%\| C \mathbin{/} B \|^{x_0,\ldots,x_n} = \forall x_{n+1},\ldots,
%x_{n+m} \| B \|^{x_n,x_{n+1},\ldots,x_{n+m}} \multimap \| C
%\|^{x_0,\ldots,x_{n-1},x_{n+1},\ldots, x_{n+m}}
%$$

Similarly, it is easy to verify that both $A
\mathbin{\downl} C$ (Equation~\ref{eq:downl} with $n=2$, $m=1$, remember that $x_2,\ldots
x_{n-1} \equiv x_2,\ldots,x_1$
and therefore equal to the empty sequence and that
$x_{n+1},\ldots,x_{n+m} \equiv x_3,\ldots, x_3 \equiv x_3$) and $A \mathbin{\downr} C$
(Equation~\ref{eq:downr} with $n=1$, $m=2$) produce the following translation for D formulas with at most two string tuples.

$$
\| A \mathbin{\downarrow} C \|^{x_1, x_2} = \forall x_0 ,x_{3}  \| A \|^{x_0,x_1,x_2,x_3} \multimap \|C \|^{x_0,x_3}
$$

%Note: $x_2$ doesn't necessarily exist 

%With $n$ is $2s(B)+1$ and $m$ is $n+2s(C)$.
%With $n$ is $2s(C\mathbin{/} B)+1$ and $m$ is $2s(B)+1$.

%$$
%\| A \mathbin{\backslash} C \|^{x_n, \ldots, x_{n+m}} = \forall x_0 ,
%\ldots, x_{n-1} \| A \|^{x_0,\ldots,x_n} \multimap \|C \|^{x_0,\ldots,x_{n-1},x_{n+1},\ldots,x_{n+m}}
%$$

%Here, $n$ is $2s(A)+1$ and $m$ is $n+2s(C)$.
%Here, $n$ is $2s(A)+1$ and $m$ is $2s(C)+1$.
%Here, $n$ is $2s(A)+1$ and $m$ is $2s(A\mathbin{\backslash}C)+1$.

In the Lambek calculus, all sorts are zero, therefore instantiating
Equation~\ref{eq:dl} with n=1, m=1 produces the following

$$
\| A \mathbin{\backslash} C \|^{x_1, x_{2}} = \forall x_0  \| A \|^{x_0,x_1} \multimap \|C \|^{x_0,x_{2}}
$$

\noindent and therefore has the translation of \cite{mill1} as a
special case.

\subsection{Examples}
\label{sec:examples}

As an illustration, let's look at the formula unfolding of $((vp
\uparrow vp) / vp)\backslash (vp\uparrow vp)$, which is the formula for ``did'' assigned to sentences like

\ex.\label{sent:did} John slept before Mary did.

\noindent by \cite{mv11displacement}. This lexical entry for
``did'' is of sort 0 and therefore has two string positions (I use 4 and 5) to
start off its translation. However, since both direct subformulas are of sort 1
its these subformulas have four position variables each. Applying the
translation for $\backslash$ shown in Equation~\ref{eq:dl} with $n=3$ (= $2s((vp\uparrow vp)/vp)+1$), $m=1$
(the sort of the complete formula being 0) gives us the following partial translation.

$$
\forall x_0 x_1 x_2 \| (vp\uparrow vp)/vp \|^{x_0,x_1,x_2,4} \multimap
\| vp \uparrow vp \|^{x_0,x_1,x_2,5}
$$

I first translate the leftmost subformula, which is of sort 1, and
apply the $/$ rule (Equation~\ref{eq:dr}) with $n = 3$ (= $2s((vp\uparrow vp)/vp)+1$) and $m=1$ (= $2s(vp)+1$) giving the following partial translation.

$$
\forall x_0 x_1 x_2 [  \forall x_3   [ \|vp\|^{4,x_3} \multimap \|
vp\uparrow vp \|^{x_0,x_1,x_2,x_3} ] \multimap
\| vp \uparrow vp \|^{x_0,x_1,x_2,5} ]
$$

Applying the translation rule for $C \uparrow B$
(Equation~\ref{eq:upl}) twice produces.

$$
\forall x_0 x_1 x_2 [  \forall x_3  [ \|vp\|^{4,x_3} \multimap
 \| vp\|^{x_1,x_2} \multimap  \| vp \|^{x_0,x_3} ] \multimap
\| vp \|^{x_1,x_2} \multimap \| vp \|^{x_0,5} ]
$$

Appendix~\ref{ex:did} gives a step-by-step derivation of the proof net for sentence~\ref{sent:did} using the contractions of Section~\ref{sec:contract} (see also Appendix~\ref{app:did} for a slightly
abbreviated alternative proof net derivation). The intelligent backtracking solution of Section~\ref{sec:eager} and \cite{moot07filter} guarantee that at each step of the computation we can make a deterministic choice for literal selection, though the reader is invited to try and find a proof by hand to convince himself that this is by no means a trivial example!

As a slightly more complicated example translation, which depends on the distinction between left wrap and right wrap, \citeasnoun{mv11displacement} give the
following formula for an object reflexive.

$$
((vp \upl np) \upr np) \downr (vp \upl np)
$$

Translating the $\downr$ connective, with input positions 3 and 4
using Equation~\ref{eq:downr} with $n=3$ (since $s((vp \upl np)
\upr np) = 2$) and $m=2$ gives the following partial translation.

$$
\forall x_0,x_1,x_2,x_5 \| (vp \upl np) \upr np
\|^{x_0,x_1,x_2,3,4,x_5} \multimap \| vp \upl np \|^{x_0,x_1,x_2,x_5}
$$

Translating the $\upr$ connective using Equation~\ref{eq:upr} with
$n=3$ and $m=2$ gives.

$$
\forall x_0,x_1,x_2,x_5 [ \| np \|^{3,4} \multimap \| vp \upl np
\|^{x_0,x_1,x_2,x_5} ] \multimap \| vp \upl np \|^{x_0,x_1,x_2,x_5}
$$

Finally, unfolding the two $\upl$ connectives (using
Equation~\ref{eq:upl}) gives.

$$
\forall x_0,x_1,x_2,x_5 [  np(3,4) \multimap np(x_1,x_2)
\multimap \| vp \|^{x_0,x_5} ] \multimap np(x_1,x_2) \multimap
\| vp \|^{x_0,x_5}
$$

Indicating that an object reflexive takes a ditransitive verb (with an
object argument spanning the positions of the reflexive) to produce a
transitive verb, which corresponds to the intuitive meaning of the lexical entry.

\subsection{Synthetic connectives}
\label{sec:synth}

\citeasnoun{mv11displacement} introduce 
the synthetic connectives --- though note that from the point of view of MILL1, \emph{all} D-connectives a synthetic MILL1-connectives, see Section~\ref{sec:correct}. These connectives can be seen as abbreviations of combinations
of a connective and an identity element ($I$ denoting the empty string
$\epsilon$ and $J$ denoting $\epsilon,\epsilon$) :

\begin{align*}
\splitc A &=_{\textit{def}} A
\uparrow I  & \textrm{Split} \\
\bridge A & =_{\textit{def}} A \odot I  & \textrm{Bridge} \\
\triangleright^{-1} A & =_{\textit{def}} J \mathbin{\backslash} A  & \textrm{Right projection} \\
\triangleright A & =_{\textit{def}} J \mathbin{\bullet} A & \textrm{Right injection} \\
\triangleleft^{-1} A & =_{\textit{def}} A \mathbin{/} J & \textrm{Left projection} \\
\triangleleft A & =_{\textit{def}} A \mathbin{\bullet} J & \textrm{Left injection} \\
\end{align*}

Figures~\ref{fig:bridge} and \ref{fig:rpi} show the proof rules for leftmost bridge/split ans right projection/injection (the
proof rules for left projection and injection as well as the proof rules for rightmost bridge and split are symmetric).

\begin{figure}

$$
\begin{array}{ccc}
\infer[\splitc E]{st:A}{s,t:\splitc A} &&
\infer[\splitc I]{s,t:\splitc A}{st:A}
\\
\\
\infer[\bridge E]{u[t]:C}{t:\bridge A & \infer*{u[st']:C}{s,t':A}}
&&
\infer[\bridge I]{st:\bridge A}{s,t:A}
\end{array}
$$

\caption{Proof rules --- leftmost split, bridge}
\label{fig:bridge}
\end{figure}

\begin{figure}

$$
\begin{array}{ccc}
\infer[\triangleright^{-1} E]{\epsilon,t:A}{t:\triangleright^{-1} A} &&
\infer[\triangleright^{-1} I]{\epsilon,t:\triangleright^{-1} A}{t:A}
\\
\\
\infer[\triangleright E]{uvu':C}{v:\triangleright A & \infer*{u,tu':C}{t:A}}
&&
\infer[\triangleright I]{\epsilon,t:\triangleright A}{t:A}
\end{array}
$$

\caption{Proof rules --- right projection, injection}
\label{fig:rpi}
\end{figure}

\editout{
The proof rule for $\splitc I$ is qualitatively different from the
corresponding $\uparrow I$ rule: the $\uparrow I$ rule hypothesizes a
$B$ formula (with a string $u$ which is unique to the proof) and uses
this formula to derive a $C$ formula with string $sut$, that is, a
string consisting of an initial, simple string $s$ followed by the
unique string tuple $u$ followed by another string tuple $t$. This
means that the $B$ hypothesis and its label uniquely determine the
conclusion string $s,t$. On the other hand, the $\splitc I$ rule
hypothesizes $I$ with the empty string }

The synthetic connectives are translated as follows (only the leftmost
split and wedge are shown, the rightmost versions are
symmetric in the variables):

\begin{align}
\| \splitc A \|^{x_0,x_1,x_1,x_2,\ldots,x_n} &=\qquad\! \| A
\|^{x_0,x_2,\ldots,x_n} \\
\| \bridge A \|^{x_0,x_2,\ldots,x_n} &= \exists x_1. \| A \|^{x_0,x_1,x_1,x_2,\ldots,x_n}\\
\| \triangleright A \|^{x_0,x_0,x_1,\ldots,x_n} &=\qquad\!  \| A
\|^{x_1,\ldots,x_n} \\
\| \triangleright^{-1} A \|^{x_1,\ldots,x_n} &= \forall x_0.  \| A \|^{x_0,x_0,x_1,\ldots,x_n}\\
\| \triangleleft A \|^{x_0,\ldots,x_n,x_{n+1},x_{n+1}} &=\qquad\ \ \, \,   \| A \|^{x_0,\ldots,x_n}\\
\| \triangleleft^{-1} A \|^{x_0,\ldots,x_n} &= \forall x_{n+1}. \| A \|^{x_0,\ldots,x_n,x_{n+1},x_{n+1}}
\end{align}

In \cite{mv11displacement}, the bridge connective appears exclusively in (positive) contexts $\bridge
(A \uparrow B)$ where it translates as.

\begin{align*}
\| \bridge( A \uparrow B) \|^{x_0,x_2}  &= \exists x_1. \| A
\uparrow B \|^{x_0,x_1,x_1,x_2}\\ &= \exists x_1. [ \| B \|^{x_1,x_1}
\multimap \| A \|^{x_0,x_2} ]
\end{align*}

The resulting formula indicates that it takes a $B$ argument spanning the empty string (anywhere) to
produce an $A$ covering the original string position $x_0$ and
$x_2$. Intuitively, this formalizes (in positive contexts) an $A$
constituent with a $B$ trace. The final translation is positive
subformula of the extraction type used in \cite{mill1}.

The split connective ($\splitc$, but also $\triangleleft$ and $\triangleright$) is more delicate, since it identifies string
position variables. This can force the identification of variables, which means
that direct application of the translation above can produce formulas
which have ``vacuous'' quantifications, though this is not harmful
(and these are easily removed in a post-processing step if
desired). However, this identification of variables means that the grammars are no longer necessarily simple in the input string as discussed in Section~\ref{sec:pospairs}. As an example,  unfolding the formula below (which is patterned after the formula for ``unfortunately'' from \cite{mv11displacement}) with input variables $x_i$ and $x_j$ forces us to identify $x_i$ and
$x_j$ as shown below, hence producing a self-loop in the input FSA.

\begin{align*}
\| \splitc A \downarrow B \|^{x_i,x_i} &= \forall x_0,x_2. \| \splitc A
\|^{x_0,x_i,x_i,x_2} \multimap \| B \|^{x_0,x_2}\\
 & = \forall x_0,x_2.   \| A \|^{x_0,x_2} \multimap \|B \|^{x_0,x_2} 
\end{align*}

Intuitively, this translation indicates
that a formula of the form $\splitc A \downarrow B$ takes its $A$
argument at any span of the string and produces a $B$ at the same position,
with the complete formula spanning the empty string. It is, in essence, a
translation of the (commutative) linear logic or LP implication into
our current context. The MIX language can easily be
generated using this property \cite{mv10d}.

It is easy to produce a complex formulas which, together with its argument, produce a complex cycle. The following formula spans the empty string after it combines with its $C$ argument.

\begin{align*}
\| (\splitc A \downarrow B)/C \|^{x_i,x_j} &= \forall x_1
 \| C \|^{x_j,x_1} \multimap \| \splitc A \downarrow B \|^{x_i,x_1}\\
 & = \forall x_1 [ \| C \|^{x_j,x_1} \multimap \forall x_0,x_2. [ \|
 \splitc A
 \|^{x_0,x_i,x_1,x_2} \multimap \| B \|^{x_0,x_2} ] ]\\
 & = \forall x_1 [ \| C \|^{x_j,x_i} \multimap \forall x_0,x_2. [ \| A
 \|^{x_0,x_2} \multimap \| B \|^{x_0,x_2} ] ]\\
& =  \| C \|^{x_j,x_i} \multimap \forall x_0,x_2. [ \| A
 \|^{x_0,x_2} \multimap \| B \|^{x_0,x_2}]
\end{align*}

The final line in the equation simply removes the $x_1$ quantifier.
Since there are no longer any occurrences of the $x_1$ variable in the
rest of the formula, this produces the equivalent formula shown. The
translation specifies that the preposition, which spans positions
$x_i$ to $x_j$ takes an $np$ argument spanning positions $x_j$ to
$x_i$, ie.\ the rightmost position of the $np$ argument is the
leftmost position of the preposition. 
%Though this again corresponds to
%the intuitive meaning for this formula, allowing these kinds of formulas would
%complicate the mapping from the input string to MILL1 formulas: it
%will become possible for the result of a concatenation of two
%non-empty strings to be the empty string, which stretches our
%intuitive notion of concatenation (and complicates our proofs!).

If we want a displacement grammar to be simple in the input string, we can restrict the
synthetic connectives used for its lexical entries to $\bridge$,  $\triangleright^{-1}$ and
$\triangleleft^{-1}$; in addition, no formulas contain the units $I$
and $J$ except where these occurrences are instances of the allowed
synthetic connectives.\footnote{Alternatively, we can allow the $\splitc$,
$\triangleright$ and $\triangleleft$ connectives but restrict them to
cases where there is strict identity of the two string positions
(disallowing instantiation of variables to obtain identity). Note that this means that formulas
of the form $\splitc A \downarrow B$ are valid only in contexts
spanning the empty string.} The only proposed lexical entries which
are not simple in this sense are those of the MIX grammar and the type for
``supposedly'' discussed above. 
%In the discussion, I will argue that this restriction
%is sensible, . Though I think this restriction can be dropped, it will
%complicate the discussion to do so, but let me give a quick sketch as
%to what the resulting system would look like.
%In the simple system, the positions variables form a series of
%(partially overlapping) paths, with each string component representing
%one of these paths. The $\bridge$,  $\triangleright^{-1}$ and
%$\triangleleft^{-1}$ connectives can be used to introduce self-loops
%for fresh variables, corresponding to the empty string $\epsilon$. However, the $\splitc$,
%$\triangleright$ and $\triangleleft$ connectives identify known
%variables, creating (possibly complex) loops spanning $\epsilon$.

The $\triangleright^{-1}$ and $\triangleleft^{-1}$ connectives,
together with atomic formulas of sort greater than 0, allow
us to encode MCFG-style analyses, as we have seen them in Section~\ref{sec:mcfg}, into D.
As an example, let's look at the unfolding of ``lezen'' which is
assigned formula $\triangleright^{-1} np\mathbin{\backslash}
(np\mathbin{\backslash} si)$ and assume ``lezen'' occupies the string
position 4,5.

$$
\forall x_2 \| np\mathbin{\backslash}
(np\mathbin{\backslash} si) \|^{x_2,x_2,4,5}
$$

Given that $s(np) = 0$ this reduces further to.

$$
\forall x_2 \forall x_1 \| np \|^{x_1,x_2} \multimap
\| np\mathbin{\backslash} si \|^{x_1,x_2,4,5}
$$

If we combine this entry with ``boeken'' from positions 1,2 (ie.\ the
formula $np(1,2)$, instantiating $x_1$ to 1 and $x_2$ to 2, this
gives the following partial translation for ``boeken lezen''

$$
\| np\mathbin{\backslash} si \|^{1,2,4,5}
$$

Similarly, ``kunnen'' with formula $\triangleright^{-1}
(np\mathbin{\backslash} si) \downarrow
(np\mathbin{\backslash} si)$ reduces as follows when occupying string
position 3,4.

$$
\forall x_2 \| (np\mathbin{\backslash} si) \downarrow
(np\mathbin{\backslash} si) \|^{x_2,x_2,3,4}
$$

Which unfolds further as.

$$
\forall x_2 \forall x_0 \forall x_5 \| np\mathbin{\backslash} si \|^{x_0,x_2,4,x_5} \multimap \|
np\mathbin{\backslash} si \|^{x_0,x_2,3,x_5}
$$

This combines with the previous translation of ``boeken lezen'',
instantiating $x_2$ to 2, $x_0$ to 1 and $x_5$ to 5, giving ``boeken
kunnen lezen'' with translation $\| np\mathbin{\backslash} si  \|^{1,2,3,5}$.

Finally, the tensed verb ``wil'' with formula $(np\mathbin{\backslash} si) \downarrow
(np\mathbin{\backslash} s)$ unfolds at position 2,3 as.

$$
\forall x_0 \forall x_3 \| np\mathbin{\backslash} si \|^{x_0,2,3,x_3}
\multimap \| np\mathbin{\backslash} s \|^{x_0,x_3}
$$

Instantiating $x_0$ to 1 and $x_3$ to 5 and combining this with the
previously computed translation of ``boeken kunnen lezen'' produces
$\| np\mathbin{\backslash} s \|^{1,5}$ for ``boeken wil kunnen lezen''. Appendix~\ref{app:nijl} shows a proof net derivation of the slightly more complex ``(dat) Jan Henk Cecilia de nijlpaarden zag helpen voeren''. Note that the axiom linkings are again fully deterministic.

\editout{
Residuation should allow us to prove:

\begin{align*}
 \bridge\, \splitc A & \vdash A \vdash \splitc\, \bridge A \\
 \triangleright  \triangleright^{-1}  A & \vdash A \vdash \triangleright^{-1}  \triangleright  A \\
\triangleleft  \triangleleft^{-1}  A & \vdash A \vdash \triangleleft^{-1}  \triangleleft  A \\
\end{align*}
}

\section{Correctness of the translation}
\label{sec:correct}

The basic idea of the correctness proof is again very simple: we use the property of focused proof search and of ludics that combinations of synchronous connectives can always be seen as instances of a synthetic synchronous connective, whereas the same holds for the asynchronous connectives. Since the translations either use a combination of $\exists$ and $\otimes$ (both synchronous) or a combination of $\forall$ and $\multimap$ (both asynchronous), it follows immediately that we can treat these combinations as synthetic connectives, giving a rule to (synthetic) rule translation.

\begin{lemma}\label{lem:sound} For every proof of $t_1:A_1,\ldots,t_n:A_n \vdash t:C$ in D, there is a proof of its translation in MILL1\end{lemma}

\paragraph{Proof}

Refer back to Figure~\ref{fig:lwrap} to see the correspondence between pairs of string
positions and tuples of strings more clearly. The rules are simply the
translation of the natural deduction rules of D, where the string
tuples have been replaced by pairs of string positions.

For the case of $\backslash E$ we are in the following situation
(let $i = \frac{1}{2}(n-1)$, $j= \frac{1}{2}(m-1)$, then $x_0,\ldots,x_n$ corresponds to a
$i$-tuple $t$, $x_n,\ldots,x_{n+m}$ to a
$j$-tuple $u$ and
$x_0,\ldots,x_{n-1},x_{n+1},\ldots,x_{n+m}$ to their concatenation
$tu$). The translations of $A$ and $A \mathbin{\backslash} C$ share
point $x_n$ and we can instantiate the universally quantified
variables of the other points of $A$ ($x_0$ to $x_{n-1}$) applying the
$\forall E$ rule $n$ times ($/$ is symmetric).

$$
\infer[\multimap E]{ \| C
  \|^{x_0,\ldots,x_{n-1},x_{n+1},\ldots,x_{n+m}}}{\|A
  \|^{x_0,\ldots,x_n} & \infer[\forall E\ \text{($n$ times)}]{ \|A \|^{x_0,\ldots,x_n} \multimap \|
  C\|^{x_0,\ldots,x_{n-1},x_{n+1},\ldots,x_{n+m}}}{\infer[=_{\textit{def}}]{\forall
  y_0,\ldots,y_{n-1} \|A \|^{y_0,\ldots,y_{n-1},x_n} \multimap \|
  C\|^{y_0,\ldots,y_{n-1},x_{n+1},\ldots,x_{n+m}}}{\| A \mathbin{\backslash}
  C \|^{x_n,\ldots,x_{n+m}}}}}
$$

For the introduction rule, we again set $i$ to $\frac{1}{2}(n-1)$ and
$j$ to $\frac{1}{2}(m-1)$, making $x_0,\ldots,x_n$ corresponds to a
$i$-tuple $t$, $x_n,\ldots,x_{n+m}$ to a
$j$-tuple $u$ and
$x_0,\ldots,x_{n-1},x_{n+1},\ldots,x_{n+m}$ to their concatenation
$tu$. In this case, induction hypothesis gives us a MILL1 proof
corresponding to  $\Gamma,t:A
\vdash tu:C$. To extend this proof to a MILL1 proof corresponding to
$\Gamma \vdash u:A \mathbin{\backslash} C$ ($/$ is again symmetric).
\editout{
  , we first need to observe that $tu$ is the complete string and
  that the only string position variable shared between $tu$
  (corresponding to $C$) and $t$ (corresponding to A) is the position
  $x_n$, which means that for all $x_i$ such that $i < n$, $x_i$ does
  not appear in $\Gamma$} which can be extended to a proof for $\Gamma,t:A
\vdash tu:C$ as follows.

$$
\infer[=_{\textit{def}}]{\| A \mathbin{\backslash}
  C\|^{x_n,\ldots,x_{x+m}}}{\infer[\forall I\ \text{($n$ times)}]{\forall
  x_0,\ldots,x_{n-1} \|A \|^{x_0,\ldots,x_n} \multimap \|
  C\|^{x_0,\ldots,x_{n-1},x_{n+1},\ldots,x_{n+m}}}{\infer[\multimap I_i]{\|A \|^{x_0,\ldots,x_n} \multimap \|
  C\|^{x_0,\ldots,x_{n-1},x_{n+1},\ldots,x_{n+m}}}{\infer*{ \|
  C\|^{x_0,\ldots,x_{n-1},x_{n+1},\ldots,x_{n+m}}}{[ \|A
    \|^{x_0,\ldots,x_n} ]^i  \; \ldots\; \Gamma}}}}
$$

The cases for $\upl$ are shown below ($\downl$ is easily verified).

$$
\infer[\multimap E]{ \| C
  \|^{x_0,x_2,\ldots,x_{n-1},x_{n+1},\ldots,x_{n+m}}}{
\infer[\forall E\ \text{($n-2$ times)}]{
 \| B \|^{x_1,\ldots,x_n} \multimap \| C \|^{x_0,x_2,\ldots,x_{n-1},x_{n+1},\ldots,x_{n+m}}   
}{\infer[=_{\textit{def}}]{\forall y_2,\ldots,y_{n-1} \| B \|^{x_1,y_2,\ldots,y_{n-1},x_n} \multimap \| C \|^{x_0,y_2,\ldots,y_{n-1},x_{n+1},\ldots,x_{n+m}}  
}{\| C \upl B \|^{x_0,x_1,x_n,\ldots,x_{n+m}}}
}
&
\| B \|^{x_1,\ldots,x_n}
}
$$

$$
\infer[=_{\textit{def}}]{\| C \mathbin{\upl}
  B\|^{x_0,x_2,\ldots,x_{n-1},x_{n+1},\ldots,x_{x+m}}}{\infer[\forall I\ \text{($n-2$ times)}]{\forall
  x_2,\ldots,x_{n-1} \|B \|^{x_1,\ldots,x_n} \multimap \|
  C\|^{x_0,x_2,\ldots,x_{n-1},x_{n+1},\ldots,x_{n+m}}}{\infer[\multimap I_i]{\|B \|^{x_1,\ldots,x_n} \multimap \|
  C\|^{x_0,x_2,\ldots,x_{n-1},x_{n+1},\ldots,x_{n+m}}}{\infer*{ \|
  C\|^{x_0,x_2,\ldots,x_{n-1},x_{n+1},\ldots,x_{n+m}}}{[ \|B
    \|^{x_1,\ldots,x_n} ]^i  \; \ldots\; \Gamma}}}}
$$

Finally, the cases for $\prodl$ are as follows.

$$
\infer[\exists E_i\ \text{twice}]{C}{\infer[=_{\textit{def}}]{\exists x_1 \exists x_n \| A \|^{x_0,\ldots,x_n} \otimes \|
  B \|^{x_n,\ldots,x_{n+m}}}{\|A \prodl B
  \|^{x_0,x_2,\ldots,x_{n-1},x_{n+1},\ldots,x_{n+m}}} & \infer[\otimes E_j]{C}{[ \| A \|^{x_0,x_1,x_n,\ldots,x_{n+m}} \otimes \|
  B \|^{x_1,\ldots,x_n}]^i & \infer*{C}{[ \| A
    \|^{x_0,x_1,x_n,\ldots,x_{n+m}}]^j  & [ \|
  B \|^{x_1,\ldots,x_n} ]^j }}}
$$

$$
\infer[=_{\textit{def}}]{ \| A \mathbin{\prodl} B \|^{x_0,x_2,\ldots,x_{n-1},x_{n+1},\ldots,x_{n+m}}}{
\infer[\exists I]{\exists x_1 \exists x_n \| A \|^{x_0,x_1,x_n\ldots,x_{n+m}} \otimes \|
  B \|^{x_1,\ldots,x_{n}}}{
\infer[\exists I]{\exists x_n \| A \|^{x_0,x_1,x_n\ldots,x_{n+m}} \otimes \|
  B \|^{x_1,\ldots,x_{n}}}{
\infer[\otimes I]{\| A \|^{x_0,x_1,x_n,\ldots,x_{n+m}} \otimes \|
  B \|^{x_1,\ldots,x_n}}{ \| A \|^{x_0,x_1,x_n,\ldots,x_{n+m}} & \|
  B \|^{x_1,\ldots,x_n}}}}}
$$

\hfill \qed

\begin{lemma}\label{lem:complete}
If the translation of a D sequent $t_1:A_1,\ldots,t_n:A_n \vdash t:C$ is provable, then there is a D
proof of $t_1:A_1,\ldots,t_n:A_n \vdash t:C$. 
\end{lemma}

%For this direction, we use the following observation (see Retor\'{e}, \cite{mr12lcg} and \cite{focus}): the par links and the universal links (the dotted logical links, called the \emph{asynchronous} connectives/links by Andreoli) as well as the tensor links and the existential links (the logical links draw with normal lines, called \emph{synchronous} links/connectives) naturally group together. When removing an asynchronous link from a proof net, we can always
%remove a number of connected par links at the same time. That is, if
%removing a par/universal link from a proof net produces a new terminal par/universal link,
%we can remove these links at the same time. Similarly, we can always remove a connected subtree of tensor/existential link to produce a number of subnets of a %proofnet. In other words,
%there is a \emph{hereditary} splitting tensor link \cite[see]{mr12lcg}.  \citeasnoun{mill1} use this same observation. 
%Finally, remark that the translation function always produces either a group of synchronous or a group of asynchronous connectives at each step.

\paragraph{Proof} This is most easily shown using proof nets, using induction on the number of links while removing them in groups of synchronous or asynchronous links corresponding to a D connective.

If there are terminal asynchronous links, then we proceed by case analysis knowing that we are dealing the result of the translation of D formulas.

The case for $C\upl B$ looks as follows.

\begin{tikzpicture}
\node (cb) at (6em,-1em) {$\| C \upl B
  \|^{x_0,x_1,x_n,\ldots,x_{n+m}}$};
\node (fx2) at (6em,2em) {$\overset{+}{\forall x_2}$};
\node (dots) at (6em,4.5em) {$\ldots$};
\node (fxn1) at (6em,7em) {$\overset{+}{\forall x_{n-1}}$};
\node (lolli) at (6em,10em) {$\overset{+}{\multimap}$};
\draw [dashed] (fx2) -- (dots);
\draw [dashed] (dots) -- (fxn1);
\draw [dashed] (fxn1) -- (lolli);
\node (b) at (1em,13em) {$\overset{-}{\| B \|^{x_1,\ldots,x_n}}$};
\node (c) at (11em,13em){$\overset{+}{\| C
    \|^{x_0,x_2,\ldots,x_{n-1},x_{n+1},\ldots,x_{n+m}}}$};
\node (gamma) at (-7em,12.75em){$\Gamma$};
\draw [dashed] (lolli) -- (b);
\draw [dashed] (lolli) -- (c);
\draw [rounded corners,fill=blue!30] (-7.5em,14em) rectangle (11.5em,18em); 
\node (pi1) at (2.0em,16em) {$\Pi$};
\end{tikzpicture}

Given that removing the portrayed links produces a proof net $\Pi$ of $\Gamma, B \vdash C$, we can apply the induction hypothesis, which gives a proof $\delta$ of
$\Gamma,u:B \vdash sut:C$, which we can extend as follows.

$$
\infer[\upl I]{s,t:C\upl B}{\infer*[\delta]{C:sut}{\Gamma & u:B}}
$$

Similarly, the par case for $\prodl$ looks as follows.

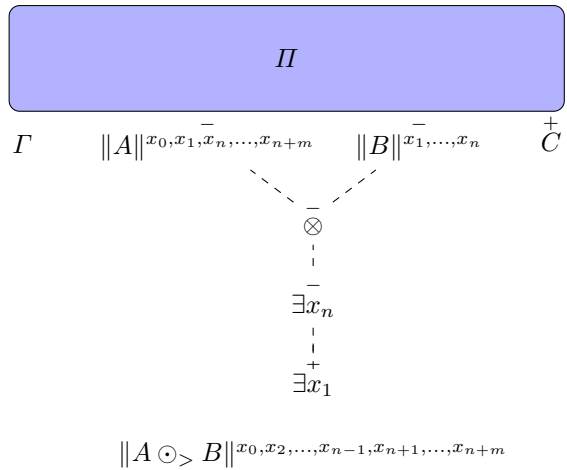
\begin{tikzpicture}
\node (cb) at (6em,1em) {$\| A \prodl B
  \|^{x_0,x_2,\ldots,x_{n-1},x_{n+1},\ldots,x_{n+m}}$};
\node (fx2) at (6em,4em) {$\overset{-}{\exists x_1}$};
\node (fxn1) at (6em,7em) {$\overset{-}{\exists x_n}$};
\node (lolli) at (6em,10em) {$\overset{-}{\otimes}$};
\draw [dashed] (fx2) -- (dots);
\draw [dashed] (dots) -- (fxn1);
\draw [dashed] (fxn1) -- (lolli);
\node (b) at (2em,13em) {$\overset{-}{\| A \|^{x_0,x_1,x_n,\ldots,x_{n+m}}}$};
\node (c) at (10em,13em){$\overset{-}{\| B
    \|^{x_1,\ldots,x_n}}$};
\node (gamma) at (-5em,12.75em){$\Gamma$};
\node (concl) at (15em,13.2em){$\overset{+}{C}$};
\draw [dashed] (lolli) -- (b);
\draw [dashed] (lolli) -- (c);
\draw [rounded corners,fill=blue!30] (-5.5em,14em) rectangle (15.5em,18em); 
\node (pi1) at (5.0em,16em) {$\Pi$};
\end{tikzpicture}

Again, we know by induction hypothesis that there is a proof $\delta$
of $\Gamma, s,t:A, u:B \vdash v[sut]:C$ and we need to show that there
is a proof of $\Gamma, sut:A\prodl B \vdash v[sut]:C$, which we do as follows.

$$
\infer[\prodl E^i]{v[sut]:C}{sut:A\prodl B &
  \infer*[\delta]{v[sut]:C}{\Gamma & [s,t:A]^i & [u:B]^i}}
$$

Suppose there are no terminal asynchronous links, then we know there must be a
group of splitting synchronous links corresponding to a D connective (a series of universal links ended by a tensor link which splits the proof net into two subnets, though the synthetic connectives of Section~\ref{sec:synth} allow for a single universal link, which is splitting by definition, since after removal of the link, all premisses of the link are the conclusion of disjoint subnets), using the standard splitting tensor argument \cite{empires}. %since the unary universal links are irrelevant for the splitting tensor.

Suppose this group of splitting links is the translation of
$\upl$, then the proof net is of the following form. 
Note that the translation of $B$ corresponds to the string tuple $u$
(with $i = \frac{1}{2}n$ components), the translation of $C\upl B$ to
the string tuple $sut$ and the translation of $C$ to the string tuple $s,t$.

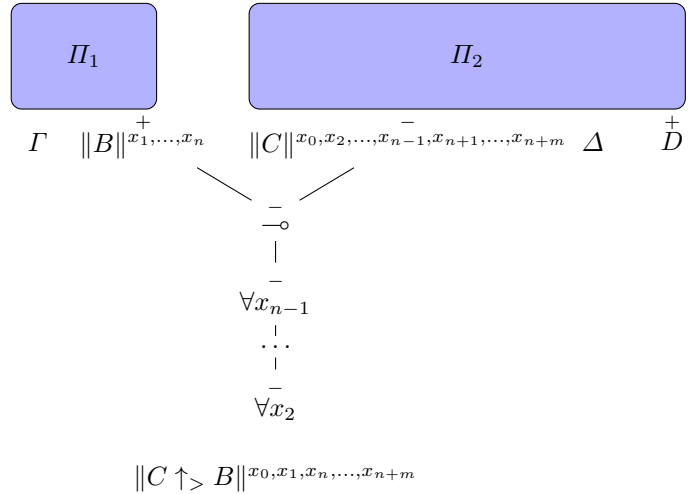
\begin{tikzpicture}
\node (cb) at (6em,0em) {$\| C \upl B
  \|^{x_0,x_1,x_n,\ldots,x_{n+m}}$};
\node (fx2) at (6em,3em) {$\overset{-}{\forall x_2}$};
\node (dots) at (6em,5em) {$\ldots$};
\node (fxn1) at (6em,7em) {$\overset{-}{\forall x_{n-1}}$};
\node (lolli) at (6em,10em) {$\overset{-}{\multimap}$};
\draw (fx2) -- (dots);
\draw (dots) -- (fxn1);
\draw (fxn1) -- (lolli);
\node (b) at (1em,13em) {$\overset{+}{\| B \|^{x_1,\ldots,x_n}}$};
\node (c) at (11em,13em){$\overset{-}{\| C
    \|^{x_0,x_2,\ldots,x_{n-1},x_{n+1},\ldots,x_{n+m}}}$};
\node (delta) at (18em,12.75em){$\Delta$};
\node (gamma) at (-3em,12.75em){$\Gamma$};
\node (d) at (21em,13.15em){$\overset{+}{D}$};
\draw (lolli) -- (b);
\draw (lolli) -- (c);
\draw [rounded corners,fill=blue!30] (-4em,14em) rectangle (1.5em,18em); 
\draw [rounded corners,fill=blue!30] (5em,14em) rectangle
(21.5em,18em); 
\node (pi1) at (-1.25em,16em) {$\Pi_1$};
\node (pi2) at (13.25em,16em) {$\Pi_2$};
\end{tikzpicture}

Therefore, we know by induction hypothesis that there is a proof
$\delta_1$ of $\Gamma \vdash u:B$ and a proof $\delta_2$ of $\Delta,
sut:C \vdash D$. We need to show that there is a proof
$\Gamma,\Delta,s,t:C\upl B \vdash D$, which we can do as follows.

$$
\infer*[\delta_2]{D}{\Delta & \infer[\upl E]{sut:C}{
\infer*[\delta_1]{u:B}{\Gamma}
&
s,t:C\upl B
}}
$$

In case the hereditary splitting tensor link is the translation of a
$\prodl$ formula, we are in the following case.

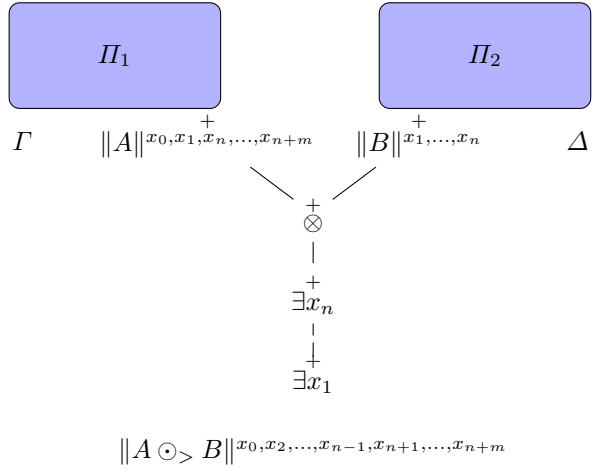
\begin{tikzpicture}
\node (cb) at (6em,1em) {$\| A \prodl B
  \|^{x_0,x_2,\ldots,x_{n-1},x_{n+1},\ldots,x_{n+m}}$};
\node (fx2) at (6em,4em) {$\overset{+}{\exists x_1}$};
\node (fxn1) at (6em,7em) {$\overset{+}{\exists x_n}$};
\node (lolli) at (6em,10em) {$\overset{+}{\otimes}$};
\draw (fx2) -- (dots);
\draw (dots) -- (fxn1);
\draw (fxn1) -- (lolli);
\node (b) at (2em,13em) {$\overset{+}{\| A \|^{x_0,x_1,x_n,\ldots,x_{n+m}}}$};
\node (c) at (10em,13em){$\overset{+}{\| B
    \|^{x_1,\ldots,x_n}}$};
\node (gamma) at (-5em,12.75em){$\Gamma$};
\node (delta) at (16em,12.75em){$\Delta$};
\draw (lolli) -- (b);
\draw (lolli) -- (c);
\draw [rounded corners,fill=blue!30] (-5.5em,14em) rectangle (2.5em,18em); 
\draw [rounded corners,fill=blue!30] (8.5em,14em) rectangle (16.5em,18em); 
\node (pi1) at (-1.5em,16em) {$\Pi_1$};
\node (pi2) at (12.5em,16em) {$\Pi_2$};
\end{tikzpicture}

Induction hypothesis gives us a proof $\delta_1$ of $\Gamma \vdash
s,t:A$ and a proof $\delta_2$ of $\Delta \vdash u:B$, which we combine
as follows.

$$
\infer[\prodl I]{sut:A\prodl B}{\infer*[\delta_1]{s,t:A}{\Gamma} & \infer*[\delta_2]{u:B}{\Delta}}
$$

\hfill \qed

\begin{theorem} Derivability of a statement in D and derivability of the translation of this statement into MILL1 coincide.

\end{theorem}

\paragraph{Proof} Immediate from Lemma~\ref{lem:sound} and \ref{lem:complete}.

The main theorem gives a simple solution to two of the main open problems from \cite{morrill2010}.

\begin{corollary} D is NP-complete
\end{corollary}

\paragraph{Proof} We have that the derivability of L, D and MILL1 are related as follows (given the translations of L and D into MILL1) $L \subset D \subset MILL1$. Therefore NP-completeness of L and MILL1 gives us NP-completeness of D.

\begin{corollary} MILL1 provides a proof net calculus for D.
\end{corollary}

\begin{corollary} D satisfies cut elimination.
\end{corollary}

Using the translation into MILL1 gives us a very easy cut elimination proof.

\section{Agreement, non-associativity and scope restrictions}
\label{sec:other}

Though I have focused only on using the first-order terms for representing string positions, I will sketch a number of other applications of the first-order terms which are orthogonal to their use for string positions, for which other extension of the Lambek calculus have introduced additional connectives and logical rules, such as the unary modalities of multimodal categorial grammar \cite{KurtoMM}.

The most obvious of these applications is for the use of linguistic features, allowing us, for example, to distinguish between nominative and accusative noun phrases $np(nom)$ and $np(acc)$ but also allowing a lexical entry to fill either role by assigning it the formula $\forall x. np(x)$.

Until now, we have only seen variables and constants as arguments of predicate symbols. When we allow more complex terms, things get more interesting. Let's only consider complex terms of the form $s(T)$ --- the well-known successor term from unary arithmetic not to be confused with the predicate symbol $s$ for sentence --- where $T$ is itself a term (complex, a variable or a constant). These complex terms allow us to implement non-associativity when we need it, using the following translation (remember that the string positions are orthogonal and can be included if needed).

$$
\begin{array}{rl}
\| A \bullet B \|^x & = \| A \|^{s(x)} \otimes \| B \|^{s(x)} \\
\| C / B \|^{s(x)} & = \| B \|^{s(x)} \multimap \| C \|^x \\
\| A \backslash C \|^{s(x)} & = \| A \|^{s(x)} \multimap \| C \|^x \\
\end{array}
$$

The translation is parametric in a single variable $x$ unique to the formula, which can get partially instantiated during the translation, producing a formula with a single free variable which is universally quantified to complete the translation. For example, a prototypical statement whose derivability presupposes associativity 

$$
a/b, b/c \vdash a/c
$$

\noindent translates as

$$
\forall x [ b(s(x)) \multimap a(x) ], \forall y [ c(s(y)) \multimap b(y) ] \vdash \forall z [ c(s(z)) \multimap a(z) ]
$$

\noindent which the reader can easily verify to be underivable. This translation generalizes both the translation of NL to MILL1 and the implementation of island constraints of \cite{mill1}.

In addition, we can handle scope restrictions in the same spirit as \cite{bernardimoot03igpl}, by translating $s_1$ as $\forall x .s(x)$, $s_2$ as $\forall x. s(s(x))$ and $s_3$ as $\forall x. s(s(s(x)))$, which are easily verified to satisfy $s_i \vdash s_j$ for $i \leq j$ and $s_i \nvdash s_j$ for $i > j$.

Scope restrictions and island constraints are some of the iconic applications of the unary modalities of multimodal categorial grammars and I consider it an attractive feature of MILL1 they permit a transparent translation of these applications.

The use of complex terms moves us rather close to the indexed grammars \cite{aho68indexed}, where complex unary term symbols play the role of a stack of indices. The linear indexed grammars \cite{gazdar88indexed} would then correspond to the restriction of quantified variables to two occurrences of opposite polarity\footnote{The encoding of non-associativity above is a clear violation of this constraint, since the quantified variable will occur in all atomic subformulas.} (or a single occurrence of any polarity; for the string position variables, they occur twice: either once as a left (resp.\ right) position of a positive atomic formula and once as a left (resp.\ right) position of a negative atomic formula or once as a left position and once as a right position of atomic formulas of the same polarity). If we restrict variables to at most two occurrences of each variable, without any restriction on the polarities, we are closer to an extension of linear indexed grammars proposed by \cite{kw95lig}, which they call partially linear PATR, and thereby closer to unification-based grammars. This restriction on quantified variables seems very interesting and naturally encompasses the restriction on string position variables. 

These are of course only suggestions, which need to be studied in more detail in future work.

\section{Conclusions and Open Questions}

First-order multiplicative intuitionistic linear logic includes several interesting subsystems: multiple context-free grammars, the Lambek calculus and the Displacement calculus. In spite of this, the computational complexity of MILL1 is the same as the complexity of the universal recognition problem for each of these individual systems. In addition, it gives a natural implementation of several additional linguistic phenomena, which would require further machinery in each of the other calculi.

MILL1 satisfies all conditions of extended Lambek calculi: it has a simple proof theory, which includes a proof net calculus, it generates the mildly context-free languages, it is NP-complete and the homomorphism for semantics consists of simply dropping the quantifiers to obtain an MILL proof --- though it is conceivable to use the first-order quantifiers for \emph{semantic} features which would have a reflection in the homomorphism. 

Many important questions have been left open. Do MILL1 grammars without complex terms (simple in the input string or not) generate exactly the MCFLs or strictly more? Do MILL1 grammars \emph{with} complex terms generate exactly the indexed languages and can we get interesting subclasses (eg.\ partially linear PATR) by restricting the variables to occur at most twice? Are there fragments of MILL1 grammars which have a polynomial fixed recognition problem? I hope these questions will receive definite answers in the future.

%What about the not-so-simple Displacement calculus? It seems that replacing string positions, as we have used them here, but states of finite state automata (as in \cite{cfgfsa}) can help us extend these results, but I feel like the simple Displacement calculus is a natural restriction which limits the calculus to the well-nested MCFGs.

%For both the Lambek calculus and MCFGs we can ensure polynomial complexity for the \emph{fixed} recognition problem: for the Lambek calculus, this is done by fixing the order of the grammar, for MCFGs by fixing ...

\bibliography{moot}
\bibliographystyle{agsm}
\vfill

\pagebreak
\appendix
\section{Example: ``John left before Mary did''}
\label{ex:did}
% Initial
\begin{tikzpicture}
\node (v8) at (14em,0em) {$v_8$};
\node (v9) at (10em,4em) {$v_9$};
\node (v10) at (8em,8em) {$v_{10}$};
\node (v11) at (12em,8em) {$v_{11}$};
\node (v17) at (18em,4em) {$v_{17}$};
\node (v18) at (16em,8em) {$v_{18}$};
\node (v19) at (20em,8em) {$v_{19}$};
\draw [dotted] (9.5em,4.7em) -- (v10);
\draw [dotted] (10.5em,4.7em) -- (v11);
\draw plot [smooth,tension=1] coordinates {(9.5em,4.7em) (10em,5.0em) (10.5em,4.7em)};
\draw [dotted] (17.5em,4.7em) -- (v18);
\draw [dotted] (18.5em,4.7em) -- (v19);
\draw plot [smooth,tension=1] coordinates {(17.5em,4.7em) (18em,5.0em) (18.5em,4.7em)};
\draw [dotted] (v8) -- (v9);
\draw [dotted] (v8) -- (v17);
\node (v12) at (10.0em,12em) {$v_{12}$};
\node (v13) at (14.0em,12em) {$v_{13}$};
\draw [dotted] (11.5em,8.7em) -- (v12);
\draw [dotted] (12.5em,8.7em) -- (v13);
\draw plot [smooth,tension=1] coordinates {(11.5em,8.7em) (12em,9.0em) (12.5em,8.7em)};
\node (v14) at (14.0em,16em) {$v_{14}$};
\draw [dotted] (v13) -- (v14);
\node (v15) at (12.0em,20em) {$v_{15}$};
\node (v16) at (16em,20em) {$v_{16}$};
\draw [dotted] (13.5em,16.7em) -- (v15);
\draw [dotted] (14.5em,16.7em) -- (v16);
\draw plot [smooth,tension=1] coordinates {(13.5em,16.7em) (14em,17.0em) (14.5em,16.7em)};
\node (v20) at (24em,0em) {$v_{20}$};
\node (v7) at (5em,0em) {$v_7$};
\node (v3) at (0em,0em) {$v_3$};
\node (v4) at (0em,4em) {$v_4$};
\draw [dotted] (v3) -- (v4);
\node (v5) at (-2em,8em) {$v_5$};
\node (v6) at (2em,8em) {$v_6$};
\draw [dotted] (-0.5em,4.7em) -- (v5);
\draw [dotted] (0.5em,4.7em) -- (v6);
\draw plot [smooth,tension=1] coordinates {(-0.5em,4.7em) (0em,5.0em) (0.5em,4.7em)};
\node (v2) at (-6em,0em) {$v_2$};
\node (v1) at (-10em,0em) {$v_1$};
\node (xp) at (0.8em,2em) {$w$};
\node (x) at (11em,2em) {$x$};
\node (z) at (17em,2em) {$z$};
\node (y) at (14.7em,14em) {$y$};
\end{tikzpicture}

$$
\begin{array}{lll|lll}
v_1 & \{ \overset{-}{np(0,1)} \} & \emptyset & v_{11} & \empty & \emptyset \\
v_2 & \{ \overset{+}{np(A,1)}, \overset{-}{s(A,2)} \} & \emptyset & v_{12} & \{ \overset{+}{np(E,Y)}, \overset{-}{s(E,Z)} \} & \emptyset \\
v_3 & \{ \overset{+}{s(3,W)}, \overset{+}{np(B,V)}, \overset{-}{s(B,W)} \} & \emptyset & v_{13} & \emptyset & \emptyset \\[3mm]
v_4 & \emptyset & \emptyset  & v_{14} & \emptyset & \emptyset \\
v_5 & \{ \overset{-}{np(w,V)} \} & \{ w \} & v_{15} & \{ \overset{-}{np(y,X)} \} & \{ y \}  \\
v_6 & \{ \overset{+}{s(w,2)} \} & \{ w \} & v_{16} & \{ \overset{+}{s(y,x)} \} & \{ x, y \} \\
v_7 & \{ \overset{-}{np(3,4)} \} & \emptyset & v_{17} & \emptyset & \emptyset  \\
v_8 & \{ \overset{-}{s(Y',5)}, \overset{+}{np(Y',X)} \} & \emptyset & v_{18} & \{ \overset{-}{np(z,Y)} \} & \{ z \} \\
v_9 & \emptyset & \emptyset & v_{19} & \{ \overset{+}{s(z,Z)} \} & \{ z \}  \\
v_{10} & \{ \overset{-}{s(X',x)}, \overset{+}{np(X',4)} \} & \{ x \} & v_{20} & \{ \overset{+}{s(0,5)} \} & \emptyset \\ 
\end{array}
$$

\pagebreak
\begin{tikzpicture}
\node (v8) at (14em,0em) {$v_8$};
\node (v9) at (10em,4em) {$v_9$};
\node (v10) at (8em,8em) {$v_{10}$};
\node (v11) at (12em,8em) {$v_{11}$};
\node (v17) at (18em,4em) {$v_{17}$};
\node (v18) at (16em,8em) {$v_{18}$};
\node (v19) at (20em,8em) {$v_{19}$};
\draw [dotted] (9.5em,4.7em) -- (v10);
\draw [dotted] (10.5em,4.7em) -- (v11);
\draw plot [smooth,tension=1] coordinates {(9.5em,4.7em) (10em,5.0em) (10.5em,4.7em)};
\draw [dotted] (17.5em,4.7em) -- (v18);
\draw [dotted] (18.5em,4.7em) -- (v19);
\draw plot [smooth,tension=1] coordinates {(17.5em,4.7em) (18em,5.0em) (18.5em,4.7em)};
\draw [dotted] (v8) -- (v9);
\draw [dotted] (v8) -- (v17);
\node (v12) at (10.0em,12em) {$v_{12}$};
\node (v13) at (14.0em,12em) {$v_{13}$};
\draw [dotted] (11.5em,8.7em) -- (v12);
\draw [dotted] (12.5em,8.7em) -- (v13);
\draw plot [smooth,tension=1] coordinates {(11.5em,8.7em) (12em,9.0em) (12.5em,8.7em)};
\node (v14) at (14.0em,16em) {$v_{14}$};
\draw [dotted] (v13) -- (v14);
\node (v15) at (12.0em,20em) {$v_{15}$};
\node (v16) at (16em,20em) {$v_{16}$};
\draw [dotted] (13.5em,16.7em) -- (v15);
\draw [dotted] (14.5em,16.7em) -- (v16);
\draw plot [smooth,tension=1] coordinates {(13.5em,16.7em) (14em,17.0em) (14.5em,16.7em)};
%\node (v20) at (24em,0em) {$v_{20}$};
\node (v7) at (5em,0em) {$v_7$};
\node (v3) at (0em,0em) {$v_3$};
\node (v4) at (0em,4em) {$v_4$};
\draw [dotted] (v3) -- (v4);
\node (v5) at (-2em,8em) {$v_5$};
\node (v6) at (2em,8em) {$v_6$};
\draw [dotted] (-0.5em,4.7em) -- (v5);
\draw [dotted] (0.5em,4.7em) -- (v6);
\draw plot [smooth,tension=1] coordinates {(-0.5em,4.7em) (0em,5.0em) (0.5em,4.7em)};
\node (v2) at (-6em,0em) {$v_2$};
\node (v1) at (-10em,0em) {$v_1$};
\node (xp) at (0.8em,2em) {$w$};
\node (x) at (11em,2em) {$x$};
\node (z) at (17em,2em) {$z$};
\node (y) at (14.7em,14em) {$y$};
\end{tikzpicture}

$$
\begin{array}{lll|lll}
v_1 & \{ \overset{-}{np(0,1)} \} & \emptyset & v_{11} & \empty & \emptyset \\
v_2 & \{ \overset{+}{np(A,1)}, \overset{-}{s(A,2)} \} & \emptyset & v_{12} & \{ \overset{+}{np(E,Y)}, \overset{-}{s(E,Z)} \} & \emptyset \\
v_3 & \{ \overset{+}{s(3,W)}, \overset{+}{np(B,V)}, \overset{-}{s(B,W)} \} & \emptyset & v_{13} & \emptyset & \emptyset \\[3mm]
v_4 & \emptyset & \emptyset  & v_{14} & \emptyset & \emptyset \\
v_5 & \{ \overset{-}{np(w,V)} \} & \{ w \} & v_{15} & \{ \overset{-}{np(y,X)} \} & \{ y \}  \\
v_6 & \{ \overset{+}{s(w,2)} \} & \{ w \} & v_{16} & \{ \overset{+}{s(y,x)} \} & \{ x, y \} \\
v_7 & \{ \overset{-}{np(3,4)} \} & \emptyset & v_{17} & \emptyset & \emptyset  \\
v_8 & \{ \overset{+}{np(0,X)} \} & \emptyset & v_{18} & \{ \overset{-}{np(z,Y)} \} & \{ z \} \\
v_9 & \emptyset & \emptyset & v_{19} & \{ \overset{+}{s(z,Z)} \} & \{ z \}  \\
v_{10} & \{ \overset{-}{s(X',x)}, \overset{+}{np(X',4)} \} & \{ x \} & %v_{20} & \{ \overset{+}{s(0,5)} \} & \emptyset 
\\ 
\end{array}
$$

\pagebreak
\begin{tikzpicture}
\node (v8) at (14em,0em) {$v_8$};
\node (v9) at (10em,4em) {$v_9$};
\node (v10) at (8em,8em) {$v_{10}$};
\node (v11) at (12em,8em) {$v_{11}$};
\node (v17) at (18em,4em) {$v_{17}$};
\node (v18) at (16em,8em) {$v_{18}$};
\node (v19) at (20em,8em) {$v_{19}$};
\draw [dotted] (9.5em,4.7em) -- (v10);
\draw [dotted] (10.5em,4.7em) -- (v11);
\draw plot [smooth,tension=1] coordinates {(9.5em,4.7em) (10em,5.0em) (10.5em,4.7em)};
\draw [dotted] (17.5em,4.7em) -- (v18);
\draw [dotted] (18.5em,4.7em) -- (v19);
\draw plot [smooth,tension=1] coordinates {(17.5em,4.7em) (18em,5.0em) (18.5em,4.7em)};
\draw [dotted] (v8) -- (v9);
\draw [dotted] (v8) -- (v17);
\node (v12) at (10.0em,12em) {$v_{12}$};
\node (v13) at (14.0em,12em) {$v_{13}$};
\draw [dotted] (11.5em,8.7em) -- (v12);
\draw [dotted] (12.5em,8.7em) -- (v13);
\draw plot [smooth,tension=1] coordinates {(11.5em,8.7em) (12em,9.0em) (12.5em,8.7em)};
\node (v14) at (14.0em,16em) {$v_{14}$};
\draw [dotted] (v13) -- (v14);
\node (v15) at (12.0em,20em) {$v_{15}$};
\node (v16) at (16em,20em) {$v_{16}$};
\draw [dotted] (13.5em,16.7em) -- (v15);
\draw [dotted] (14.5em,16.7em) -- (v16);
\draw plot [smooth,tension=1] coordinates {(13.5em,16.7em) (14em,17.0em) (14.5em,16.7em)};
%\node (v20) at (24em,0em) {$v_{20}$};
\node (v7) at (5em,0em) {$v_7$};
\node (v3) at (0em,0em) {$v_3$};
\node (v4) at (0em,4em) {$v_4$};
\draw [dotted] (v3) -- (v4);
\node (v5) at (-2em,8em) {$v_5$};
\node (v6) at (2em,8em) {$v_6$};
\draw [dotted] (-0.5em,4.7em) -- (v5);
\draw [dotted] (0.5em,4.7em) -- (v6);
\draw plot [smooth,tension=1] coordinates {(-0.5em,4.7em) (0em,5.0em) (0.5em,4.7em)};
\node (v2) at (-6em,0em) {$v_2$};
%\node (v1) at (-10em,0em) {$v_1$};
\node (xp) at (0.8em,2em) {$w$};
\node (x) at (11em,2em) {$x$};
\node (z) at (17em,2em) {$z$};
\node (y) at (14.7em,14em) {$y$};
\end{tikzpicture}

$$
\begin{array}{lll|lll}
v_2 & \{ \overset{+}{np(A,1)}, \overset{-}{s(A,2)} \} & \emptyset & v_{11} & \empty & \emptyset \\
v_3 & \{ \overset{+}{s(3,W)}, \overset{+}{np(B,V)}, \overset{-}{s(B,W)} \} & \emptyset & v_{12} & \{ \overset{+}{np(E,Y)}, \overset{-}{s(E,Z)} \} & \emptyset \\[3mm]
v_4 & \emptyset & \emptyset  & v_{13} & \emptyset & \emptyset \\
v_5 & \{ \overset{-}{np(w,V)} \} & \{ w \} &  v_{14} & \emptyset & \emptyset  \\
v_6 & \{ \overset{+}{s(w,2)} \} & \{ w \} & v_{15} & \{ \overset{-}{np(y,1)} \} & \{ y \} \\
v_7 & \{ \overset{-}{np(3,4)} \} & \emptyset &  v_{16} & \{ \overset{+}{s(y,x)} \} & \{ x, y \} \\[3mm]
v_8 & \emptyset & \emptyset & v_{17} & \emptyset & \emptyset \\
v_9 & \emptyset & \emptyset &  v_{18} & \{ \overset{-}{np(z,Y)} \} & \{ z \} \\
v_{10} & \{ \overset{-}{s(X',x)}, \overset{+}{np(X',4)} \} & \{ x \} & v_{19} & \{ \overset{+}{s(z,Z)} \} & \{ z \}
\\ 
\end{array}
$$

\pagebreak
\begin{tikzpicture}
\node (v8) at (10em,0em) {$v_8$};
\node (v9) at (10em,4em) {$v_9$};
\node (v10) at (8em,8em) {$v_{10}$};
\node (v11) at (12em,8em) {$v_{11}$};
\draw [dotted] (9.5em,4.7em) -- (v10);
\draw [dotted] (10.5em,4.7em) -- (v11);
\draw plot [smooth,tension=1] coordinates {(9.5em,4.7em) (10em,5.0em) (10.5em,4.7em)};
\draw [dotted] (v8) -- (v9);
\node (v12) at (10.0em,12em) {$v_{12}$};
\node (v13) at (14.0em,12em) {$v_{13}$};
\draw [dotted] (11.5em,8.7em) -- (v12);
\draw [dotted] (12.5em,8.7em) -- (v13);
\draw plot [smooth,tension=1] coordinates {(11.5em,8.7em) (12em,9.0em) (12.5em,8.7em)};
\node (v14) at (14.0em,16em) {$v_{14}$};
\draw [dotted] (v13) -- (v14);
\node (v15) at (12.0em,20em) {$v_{15}$};
\node (v16) at (16em,20em) {$v_{16}$};
\draw [dotted] (13.5em,16.7em) -- (v15);
\draw [dotted] (14.5em,16.7em) -- (v16);
\draw plot [smooth,tension=1] coordinates {(13.5em,16.7em) (14em,17.0em) (14.5em,16.7em)};
\node (v7) at (5em,0em) {$v_7$};
\node (v3) at (0em,0em) {$v_3$};
\node (v4) at (0em,4em) {$v_4$};
\draw [dotted] (v3) -- (v4);
\node (v5) at (-2em,8em) {$v_5$};
\node (v6) at (2em,8em) {$v_6$};
\draw [dotted] (-0.5em,4.7em) -- (v5);
\draw [dotted] (0.5em,4.7em) -- (v6);
\draw plot [smooth,tension=1] coordinates {(-0.5em,4.7em) (0em,5.0em) (0.5em,4.7em)};
\node (xp) at (0.8em,2em) {$w$};
\node (x) at (11em,2em) {$x$};
\node (y) at (14.7em,14em) {$y$};
\end{tikzpicture}

$$
\begin{array}{lll|lll}
v_3 & \{ \overset{+}{s(3,W)}, \overset{+}{np(B,V)}, \overset{-}{s(B,W)} \} & \emptyset & v_{10} & \{ \overset{-}{s(X',x)}, \overset{+}{np(X',4)} \} & \{ x \} \\[3mm]
v_4 & \emptyset & \emptyset  & v_{11} & \emptyset & \emptyset \\
v_5 & \{ \overset{-}{np(w,V)} \} & \{ w \}   & v_{12} & \{ \overset{+}{np(E,1)}, \overset{-}{s(E,2)} \} & \emptyset \\
v_6 & \{ \overset{+}{s(w,2)} \} & \{ w \}   & v_{13} & \emptyset & \emptyset  \\
v_7 & \{ \overset{-}{np(3,4)} \} & \emptyset & v_{14} & \emptyset & \emptyset  \\
v_8 & \emptyset & \emptyset & v_{15} & \{ \overset{-}{np(y,1)} \} & \{ y \}  \\
v_9 & \emptyset & \emptyset & v_{16} & \{ \overset{+}{s(y,x)} \} & \{ x, y \}
\\
\end{array}
$$

\pagebreak
\begin{tikzpicture}
\node (v8) at (10em,0em) {$v_8$};
\node (v9) at (10em,4em) {$v_9$};
\node (v10) at (8em,8em) {$v_{10}$};
\node (v11) at (12em,8em) {$v_{11}$};
\draw [dotted] (9.5em,4.7em) -- (v10);
\draw [dotted] (10.5em,4.7em) -- (v11);
\draw plot [smooth,tension=1] coordinates {(9.5em,4.7em) (10em,5.0em) (10.5em,4.7em)};
\draw [dotted] (v8) -- (v9);
\node (v12) at (10.0em,12em) {$v_{3}$};
\node (v13) at (14.0em,12em) {$v_{13}$};
\draw [dotted] (11.5em,8.7em) -- (v12);
\draw [dotted] (12.5em,8.7em) -- (v13);
\draw plot [smooth,tension=1] coordinates {(11.5em,8.7em) (12em,9.0em) (12.5em,8.7em)};
\node (v14) at (14.0em,16em) {$v_{14}$};
\draw [dotted] (v13) -- (v14);
\node (v15) at (12.0em,20em) {$v_{15}$};
\node (v16) at (16em,20em) {$v_{16}$};
\draw [dotted] (13.5em,16.7em) -- (v15);
\draw [dotted] (14.5em,16.7em) -- (v16);
\draw plot [smooth,tension=1] coordinates {(13.5em,16.7em) (14em,17.0em) (14.5em,16.7em)};
\node (v7) at (5em,0em) {$v_7$};
%\node (v3) at (0em,0em) {$v_3$};
%\node (v4) at (0em,4em) {$v_4$};
%\draw [dotted] (v3) -- (v4);
%\node (v5) at (-2em,8em) {$v_5$};
%\node (v6) at (2em,8em) {$v_6$};
%\draw [dotted] (-0.5em,4.7em) -- (v5);
%\draw [dotted] (0.5em,4.7em) -- (v6);
%\draw plot [smooth,tension=1] coordinates {(-0.5em,4.7em) (0em,5.0em) (0.5em,4.7em)};
%\node (xp) at (0.8em,2em) {$w$};
\node (x) at (11em,2em) {$x$};
\node (y) at (14.7em,14em) {$y$};
\end{tikzpicture}

$$
\begin{array}{lll|lll}
v_3 & \{ \overset{+}{s(3,W)}, \overset{+}{np(B,1)}, \overset{-}{s(B,W)} \} & \emptyset & v_{10} & \{ \overset{-}{s(X',x)}, \overset{+}{np(X',4)} \} & \{ x \} \\[3mm]
%v_4 & \emptyset & \emptyset  & 
& & & v_{11} & \emptyset & \emptyset \\
%v_5 & \{ \overset{-}{np(w,1)} \} & \{ w \}   &
& & & v_{12} & \{ \overset{+}{np(w,1)}, \overset{-}{s(w,2)} \} & \emptyset \\
%v_6 & \{ \overset{+}{s(w,2)} \} & \{ w \}   & 
& & & v_{13} & \emptyset & \emptyset  \\
v_7 & \{ \overset{-}{np(3,4)} \} & \emptyset & v_{14} & \emptyset & \emptyset  \\
v_8 & \emptyset & \emptyset & v_{15} & \{ \overset{-}{np(y,1)} \} & \{ y \}  \\
v_9 & \emptyset & \emptyset & v_{16} & \{ \overset{+}{s(y,x)} \} & \{ x, y \}
\\
\end{array}
$$

\pagebreak
\begin{tikzpicture}
\node (v8) at (10em,0em) {$v_8$};
\node (v9) at (10em,4em) {$v_9$};
\node (v10) at (8em,8em) {$v_{10}$};
\node (v11) at (12em,8em) {$v_{11}$};
\draw [dotted] (9.5em,4.7em) -- (v10);
\draw [dotted] (10.5em,4.7em) -- (v11);
\draw plot [smooth,tension=1] coordinates {(9.5em,4.7em) (10em,5.0em) (10.5em,4.7em)};
\draw [dotted] (v8) -- (v9);
\node (v12) at (12.0em,12em) {$v_{3}$};
\node (v7) at (5em,0em) {$v_7$};
\node (x) at (11em,2em) {$x$};
\draw [dotted] plot [smooth, tension=1] coordinates {(11.6em,8.5em) (11em,10em) (11.6em,11.5em)};
\draw [dotted] plot [smooth, tension=1] coordinates {(12.4em,8.5em) (13em,10em) (12.4em,11.5em)};
\draw plot [smooth,tension=1] coordinates {(11.6em,8.5em) (12em,8.7em) (12.4em,8.5em)};
\end{tikzpicture}

$$
\begin{array}{lll|lll}
v_3 & \{ \overset{+}{s(3,x)}\} & \{ x \} & 
v_{10} & \{ \overset{-}{s(X',x)}, \overset{+}{np(X',4)} \} & \{ x \} \\[3mm]
& & & v_{11} & \emptyset & \emptyset \\
& & & 
\\

& & & 
 \\
v_7 & \{ \overset{-}{np(3,4)} \} & \emptyset & 
\\
v_8 & \emptyset & \emptyset &
\\
v_9 & \emptyset & \emptyset &
\\
\end{array}
$$

\pagebreak
\begin{tikzpicture}
\node (v8) at (10em,0em) {$v_8$};
\node (v9) at (10em,4em) {$v_9$};
\node (v10) at (8em,8em) {$v_{10}$};
\node (v11) at (12em,8em) {$v_{11}$};
\draw [dotted] (9.5em,4.7em) -- (v10);
\draw [dotted] (10.5em,4.7em) -- (v11);
\draw plot [smooth,tension=1] coordinates {(9.5em,4.7em) (10em,5.0em) (10.5em,4.7em)};
\draw [dotted] (v8) -- (v9);
%\node (v12) at (12.0em,12em) {$v_{3}$};
\node (v7) at (5em,0em) {$v_7$};
\node (x) at (11em,2em) {$x$};
%
%\draw [dotted] plot [smooth, tension=1] coordinates {(11.6em,8.5em) (11em,10em) (11.6em,11.5em)};
%\draw [dotted] plot [smooth, tension=1] coordinates {(12.4em,8.5em) (13em,10em) (12.4em,11.5em)};
%\draw plot [smooth,tension=1] coordinates {(11.6em,8.5em) (12em,8.7em) (12.4em,8.5em)};
\end{tikzpicture}

$$
\begin{array}{lll|lll}
%v_3 & \{ \overset{+}{s(3,x)}\} & \{ x \} &
& & & 
v_{10} & \{ \overset{-}{s(X',x)}, \overset{+}{np(X',4)} \} & \{ x \} \\
& & & v_{11} & \{ \overset{+}{s(3,x)}\} &  \{ x \} \\
& & & 
\\

& & & 
 \\
v_7 & \{ \overset{-}{np(3,4)} \} & \emptyset & 
\\
v_8 & \emptyset & \emptyset &
\\
v_9 & \emptyset & \emptyset &
\\
\end{array}
$$

\pagebreak
\begin{tikzpicture}
\node (v8) at (10em,0em) {$v_8$};
\node (v9) at (10em,4em) {$v_9$};
\node (v10) at (10em,8em) {$v_{10}$};
%\node (v11) at (12em,8em) {$v_{11}$};
%\draw [dotted] (9.5em,4.7em) -- (v10);
%\draw [dotted] (10.5em,4.7em) -- (v11);
%\draw plot [smooth,tension=1] coordinates {(9.5em,4.7em) (10em,5.0em) (10.5em,4.7em)};
\draw [dotted] (v8) -- (v9);
%\node (v12) at (12.0em,12em) {$v_{3}$};
\node (v7) at (5em,0em) {$v_7$};
\node (x) at (11em,2em) {$x$};
\draw [dotted] plot [smooth, tension=1] coordinates {(9.6em,4.5em) (9em,6em) (9.6em,7.5em)};
\draw [dotted] plot [smooth, tension=1] coordinates {(10.4em,4.5em) (11em,6em) (10.4em,7.5em)};
\draw plot [smooth,tension=1] coordinates {(9.6em,4.5em) (10em,4.7em) (10.4em,4.5em)};
\end{tikzpicture}

$$
\begin{array}{lll|lll}
%v_3 & \{ \overset{+}{s(3,x)}\} & \{ x \} &
& & & 
v_{10} & \{ \overset{+}{np(3,4)} \} & \{ x \} \\
& & & %v_{11} & \{ \overset{+}{s(3,x)}\} &  \{ x \} 
\\
& & & 
\\

& & & 
 \\
v_7 & \{ \overset{-}{np(3,4)} \} & \emptyset & 
\\
v_8 & \emptyset & \emptyset &
\\
v_9 & \emptyset & \emptyset &
\\
\end{array}
$$

%%%% end
\section{Example: ``John left before Mary did''}
\label{app:did}

\begin{sideways}
\begin{tikzpicture}
\node (j) at (-2em,0em) {$\overset{-}{np(0,1)}$};
\node (l) at (2em,0em) {$\overset{-}{vp(1,2)}$};
\node (bf) at (10em,0em) {$\overset{-}{\multimap}$};
\node (b1) at (8em,3em) {$\overset{-}{\multimap}$};
\node (b2) at (12em,3em) {$\overset{+}{s(3,W)}$};
\node (b11) at (6em,6em) {$\overset{+}{vp(V,2)}$};
\node (b12) at (10em,6em) {$\overset{-}{vp(V,W)}$};
\draw (b1) -- (b11);
\draw (b1) -- (b12);
\draw (bf) -- (b1);
\draw (bf) -- (b2);
\node (m) at (16em,0em) {$\overset{-}{np(3,4)}$};
\node (forall) at (26.5em,3em) {$\overset{+}{\forall x}$};
\node (d11) at (22.25em,9em) {$\overset{-}{vp(4,x)}$};
\node (d111) at (20em,12em) {$\overset{+}{np(X',4)}$};
\node (d112) at (24.5em,12em) {$\overset{-}{s(X',x)}$};
\node (d121) at (28.5em,12em) {$\overset{-}{vp(Y,Z)}$};
\node (d122) at (33em,12em) {$\overset{+}{vp(X,x)}$};
\node (d12) at (30.75em,9em) {$\overset{+}{\multimap}$};
\node (d1) at (26.5em,6em) {$\overset{+}{\multimap}$};
\node (d221) at (40em,9em) {$\overset{+}{np(Y',X)}$};
\node (d222) at (45em,9em) {$\overset{-}{s(Y',5)}$};
\node (d21) at (37em,6em) {$\overset{+}{vp(Y,Z)}$};
\node (d22) at (42.5em,6em) {$\overset{-}{vp(X,5)}$};
\node (d2) at (39.75em,3em) {$\overset{-}{\multimap}$};
\node (df) at (33.125em,0em) {$\overset{-}{\multimap}$};
\draw (df) -- (forall);
\draw [dashed] (forall) -- (d1);
\draw [dashed] (d1) -- (d11);
\draw [dashed] (d1) -- (d12);
\draw [dashed] (d12) -- (d121);
\draw [dashed] (d12) -- (d122);
\draw (d11) -- (d111);
\draw (d11) -- (d112);
%\draw (df) -- (d1);
\draw (df) -- (d2);
\draw (d2) -- (d21);
\draw (d2) -- (d22);
\draw (d22) -- (d221);
\draw (d22) -- (d222);
\node (s) at (47em,0em) {$\overset{+}{s(0,5)}$};
% Axioms
\end{tikzpicture}
\end{sideways}

\begin{sideways}
\begin{tikzpicture}
\node (j) at (-2em,0em) {$\overset{-}{np(0,1)}$};
\node (l) at (2em,0em) {$\overset{-}{vp(1,2)}$};
\node (bf) at (10em,0em) {$\overset{-}{\multimap}$};
\node (b1) at (8em,3em) {$\overset{-}{\multimap}$};
\node (b2) at (12em,3em) {$\overset{+}{s(3,W)}$};
\node (b11) at (6em,6em) {$\overset{+}{vp(V,2)}$};
\node (b12) at (10em,6em) {$\overset{-}{vp(V,W)}$};
\draw (b1) -- (b11);
\draw (b1) -- (b12);
\draw (bf) -- (b1);
\draw (bf) -- (b2);
\node (m) at (16em,0em) {$\overset{-}{np(3,4)}$};
\node (forall) at (26.5em,3em) {$\overset{+}{\forall x}$};
\node (d11) at (22.25em,9em) {$\overset{-}{vp(4,x)}$};
\node (d111) at (20em,12em) {$\overset{+}{np(3,4)}$};
\node (d112) at (24.5em,12em) {$\overset{-}{s(3,x)}$};
\node (d121) at (28.5em,12em) {$\overset{-}{vp(Y,Z)}$};
\node (d122) at (33em,12em) {$\overset{+}{vp(X,x)}$};
\node (d12) at (30.75em,9em) {$\overset{+}{\multimap}$};
\node (d1) at (26.5em,6em) {$\overset{+}{\multimap}$};
\node (d221) at (40em,9em) {$\overset{+}{np(Y',X)}$};
\node (d222) at (45em,9em) {$\overset{-}{s(Y',5)}$};
\node (d21) at (37em,6em) {$\overset{+}{vp(Y,Z)}$};
\node (d22) at (42.5em,6em) {$\overset{-}{vp(X,5)}$};
\node (d2) at (39.75em,3em) {$\overset{-}{\multimap}$};
\node (df) at (33.125em,0em) {$\overset{-}{\multimap}$};
\draw (df) -- (forall);
\draw [dashed] (forall) -- (d1);
\draw [dashed] (d1) -- (d11);
\draw [dashed] (d1) -- (d12);
\draw [dashed] (d12) -- (d121);
\draw [dashed] (d12) -- (d122);
\draw (d11) -- (d111);
\draw (d11) -- (d112);
%\draw (df) -- (d1);
\draw (df) -- (d2);
\draw (d2) -- (d21);
\draw (d2) -- (d22);
\draw (d22) -- (d221);
\draw (d22) -- (d222);
\node (s) at (47em,0em) {$\overset{+}{s(0,5)}$};
% Axioms X' = 3
\draw (m) -- (16em,14em) -- (20em,14em) -- (d111);
\end{tikzpicture}
\end{sideways}

\begin{sideways}
\begin{tikzpicture}
\node (j) at (-2em,0em) {$\overset{-}{np(0,1)}$};
\node (l) at (2em,0em) {$\overset{-}{vp(1,2)}$};
\node (bf) at (10em,0em) {$\overset{-}{\multimap}$};
\node (b1) at (8em,3em) {$\overset{-}{\multimap}$};
\node (b2) at (12em,3em) {$\overset{+}{s(3,x)}$};
\node (b11) at (6em,6em) {$\overset{+}{vp(V,2)}$};
\node (b12) at (10em,6em) {$\overset{-}{vp(V,x)}$};
\draw (b1) -- (b11);
\draw (b1) -- (b12);
\draw (bf) -- (b1);
\draw (bf) -- (b2);
\node (m) at (16em,0em) {$\overset{-}{np(3,4)}$};
\node (forall) at (26.5em,3em) {$\overset{+}{\forall x}$};
\node (d11) at (22.25em,9em) {$\overset{-}{vp(4,x)}$};
\node (d111) at (20em,12em) {$\overset{+}{np(3,4)}$};
\node (d112) at (24.5em,12em) {$\overset{-}{s(3,x)}$};
\node (d121) at (28.5em,12em) {$\overset{-}{vp(Y,Z)}$};
\node (d122) at (33em,12em) {$\overset{+}{vp(X,x)}$};
\node (d12) at (30.75em,9em) {$\overset{+}{\multimap}$};
\node (d1) at (26.5em,6em) {$\overset{+}{\multimap}$};
\node (d221) at (40em,9em) {$\overset{+}{np(Y',X)}$};
\node (d222) at (45em,9em) {$\overset{-}{s(Y',5)}$};
\node (d21) at (37em,6em) {$\overset{+}{vp(Y,Z)}$};
\node (d22) at (42.5em,6em) {$\overset{-}{vp(X,5)}$};
\node (d2) at (39.75em,3em) {$\overset{-}{\multimap}$};
\node (df) at (33.125em,0em) {$\overset{-}{\multimap}$};
\draw (df) -- (forall);
\draw [dashed] (forall) -- (d1);
\draw [dashed] (d1) -- (d11);
\draw [dashed] (d1) -- (d12);
\draw [dashed] (d12) -- (d121);
\draw [dashed] (d12) -- (d122);
\draw (d11) -- (d111);
\draw (d11) -- (d112);
%\draw (df) -- (d1);
\draw (df) -- (d2);
\draw (d2) -- (d21);
\draw (d2) -- (d22);
\draw (d22) -- (d221);
\draw (d22) -- (d222);
\node (s) at (47em,0em) {$\overset{+}{s(0,5)}$};
% Axioms
% X' = 3
\draw (m) -- (16em,14em) -- (20em,14em) -- (d111);
% W = x
\draw (b2) -- (12em,15em) -- (24.5em,15em) -- (d112);
\end{tikzpicture}
\end{sideways}

\begin{sideways}
\begin{tikzpicture}
\node (j) at (-2em,0em) {$\overset{-}{np(0,1)}$};
\node (l) at (2em,0em) {$\overset{-}{vp(1,2)}$};
\node (bf) at (10em,0em) {$\overset{-}{\multimap}$};
\node (b1) at (8em,3em) {$\overset{-}{\multimap}$};
\node (b2) at (12em,3em) {$\overset{+}{s(3,x)}$};
\node (b11) at (6em,6em) {$\overset{+}{vp(V,2)}$};
\node (b12) at (10em,6em) {$\overset{-}{vp(V,x)}$};
\draw (b1) -- (b11);
\draw (b1) -- (b12);
\draw (bf) -- (b1);
\draw (bf) -- (b2);
\node (m) at (16em,0em) {$\overset{-}{np(3,4)}$};
\node (forall) at (26.5em,3em) {$\overset{+}{\forall x}$};
\node (d11) at (22.25em,9em) {$\overset{-}{vp(4,x)}$};
\node (d111) at (20em,12em) {$\overset{+}{np(3,4)}$};
\node (d112) at (24.5em,12em) {$\overset{-}{s(3,x)}$};
\node (d121) at (28.5em,12em) {$\overset{-}{vp(1,2)}$};
\node (d122) at (33em,12em) {$\overset{+}{vp(X,x)}$};
\node (d12) at (30.75em,9em) {$\overset{+}{\multimap}$};
\node (d1) at (26.5em,6em) {$\overset{+}{\multimap}$};
\node (d221) at (40em,9em) {$\overset{+}{np(Y',1)}$};
\node (d222) at (45em,9em) {$\overset{-}{s(Y',5)}$};
\node (d21) at (37em,6em) {$\overset{+}{vp(1,2)}$};
\node (d22) at (42.5em,6em) {$\overset{-}{vp(X,5)}$};
\node (d2) at (39.75em,3em) {$\overset{-}{\multimap}$};
\node (df) at (33.125em,0em) {$\overset{-}{\multimap}$};
\draw (df) -- (forall);
\draw [dashed] (forall) -- (d1);
\draw [dashed] (d1) -- (d11);
\draw [dashed] (d1) -- (d12);
\draw [dashed] (d12) -- (d121);
\draw [dashed] (d12) -- (d122);
\draw (d11) -- (d111);
\draw (d11) -- (d112);
%\draw (df) -- (d1);
\draw (df) -- (d2);
\draw (d2) -- (d21);
\draw (d2) -- (d22);
\draw (d22) -- (d221);
\draw (d22) -- (d222);
\node (s) at (47em,0em) {$\overset{+}{s(0,5)}$};
% Axioms
% X' = 3
\draw (m) -- (16em,14em) -- (20em,14em) -- (d111);
% W = x
\draw (b2) -- (12em,15em) -- (24.5em,15em) -- (d112);
% V = X
%\draw (b12) -- (10em,16em) -- (33em,16em) -- (d122);
% Y = 1, Z = 2
\draw (l) -- (2em,18em) -- (37em,18em) -- (d21);
\end{tikzpicture}
\end{sideways}

\begin{sideways}
\begin{tikzpicture}
\node (j) at (-2em,0em) {$\overset{-}{np(0,1)}$};
\node (l) at (2em,0em) {$\overset{-}{vp(1,2)}$};
\node (bf) at (10em,0em) {$\overset{-}{\multimap}$};
\node (b1) at (8em,3em) {$\overset{-}{\multimap}$};
\node (b2) at (12em,3em) {$\overset{+}{s(3,x)}$};
\node (b11) at (6em,6em) {$\overset{+}{vp(1,2)}$};
\node (b12) at (10em,6em) {$\overset{-}{vp(1,x)}$};
\draw (b1) -- (b11);
\draw (b1) -- (b12);
\draw (bf) -- (b1);
\draw (bf) -- (b2);
\node (m) at (16em,0em) {$\overset{-}{np(3,4)}$};
\node (forall) at (26.5em,3em) {$\overset{+}{\forall x}$};
\node (d11) at (22.25em,9em) {$\overset{-}{vp(4,x)}$};
\node (d111) at (20em,12em) {$\overset{+}{np(3,4)}$};
\node (d112) at (24.5em,12em) {$\overset{-}{s(3,x)}$};
\node (d121) at (28.5em,12em) {$\overset{-}{vp(1,2)}$};
\node (d122) at (33em,12em) {$\overset{+}{vp(1,x)}$};
\node (d12) at (30.75em,9em) {$\overset{+}{\multimap}$};
\node (d1) at (26.5em,6em) {$\overset{+}{\multimap}$};
\node (d221) at (40em,9em) {$\overset{+}{np(Y',1)}$};
\node (d222) at (45em,9em) {$\overset{-}{s(Y',5)}$};
\node (d21) at (37em,6em) {$\overset{+}{vp(1,2)}$};
\node (d22) at (42.5em,6em) {$\overset{-}{vp(1,5)}$};
\node (d2) at (39.75em,3em) {$\overset{-}{\multimap}$};
\node (df) at (33.125em,0em) {$\overset{-}{\multimap}$};
\draw (df) -- (forall);
\draw [dashed] (forall) -- (d1);
\draw [dashed] (d1) -- (d11);
\draw [dashed] (d1) -- (d12);
\draw [dashed] (d12) -- (d121);
\draw [dashed] (d12) -- (d122);
\draw (d11) -- (d111);
\draw (d11) -- (d112);
%\draw (df) -- (d1);
\draw (df) -- (d2);
\draw (d2) -- (d21);
\draw (d2) -- (d22);
\draw (d22) -- (d221);
\draw (d22) -- (d222);
\node (s) at (47em,0em) {$\overset{+}{s(0,5)}$};
% Axioms
% X' = 3
\draw (m) -- (16em,14em) -- (20em,14em) -- (d111);
% W = x
\draw (b2) -- (12em,15em) -- (24.5em,15em) -- (d112);
% Y = 1, Z = 2
\draw (l) -- (2em,18em) -- (37em,18em) -- (d21);
% V = X
\draw (b12) -- (10em,16em) -- (33em,16em) -- (d122);
% V = X = 1
\draw (b11) -- (6em,17em) -- (28.5em,17em) -- (d121);
\end{tikzpicture}
\end{sideways}

\begin{sideways}
\begin{tikzpicture}
\node (j) at (-2em,0em) {$\overset{-}{np(0,1)}$};
\node (l) at (2em,0em) {$\overset{-}{vp(1,2)}$};
\node (bf) at (10em,0em) {$\overset{-}{\multimap}$};
\node (b1) at (8em,3em) {$\overset{-}{\multimap}$};
\node (b2) at (12em,3em) {$\overset{+}{s(3,x)}$};
\node (b11) at (6em,6em) {$\overset{+}{vp(1,2)}$};
\node (b12) at (10em,6em) {$\overset{-}{vp(1,x)}$};
\draw (b1) -- (b11);
\draw (b1) -- (b12);
\draw (bf) -- (b1);
\draw (bf) -- (b2);
\node (m) at (16em,0em) {$\overset{-}{np(3,4)}$};
\node (forall) at (26.5em,3em) {$\overset{+}{\forall x}$};
\node (d11) at (22.25em,9em) {$\overset{-}{vp(4,x)}$};
\node (d111) at (20em,12em) {$\overset{+}{np(3,4)}$};
\node (d112) at (24.5em,12em) {$\overset{-}{s(3,x)}$};
\node (d121) at (28.5em,12em) {$\overset{-}{vp(1,2)}$};
\node (d122) at (33em,12em) {$\overset{+}{vp(1,x)}$};
\node (d12) at (30.75em,9em) {$\overset{+}{\multimap}$};
\node (d1) at (26.5em,6em) {$\overset{+}{\multimap}$};
\node (d221) at (40em,9em) {$\overset{+}{np(0,1)}$};
\node (d222) at (45em,9em) {$\overset{-}{s(0,5)}$};
\node (d21) at (37em,6em) {$\overset{+}{vp(1,2)}$};
\node (d22) at (42.5em,6em) {$\overset{-}{vp(1,5)}$};
\node (d2) at (39.75em,3em) {$\overset{-}{\multimap}$};
\node (df) at (33.125em,0em) {$\overset{-}{\multimap}$};
\draw (df) -- (forall);
\draw [dashed] (forall) -- (d1);
\draw [dashed] (d1) -- (d11);
\draw [dashed] (d1) -- (d12);
\draw [dashed] (d12) -- (d121);
\draw [dashed] (d12) -- (d122);
\draw (d11) -- (d111);
\draw (d11) -- (d112);
%\draw (df) -- (d1);
\draw (df) -- (d2);
\draw (d2) -- (d21);
\draw (d2) -- (d22);
\draw (d22) -- (d221);
\draw (d22) -- (d222);
\node (s) at (47em,0em) {$\overset{+}{s(0,5)}$};
% Axioms
% X' = 3
\draw (m) -- (16em,14em) -- (20em,14em) -- (d111);
% W = x
\draw (b2) -- (12em,15em) -- (24.5em,15em) -- (d112);
% Y = 1, Z = 2
\draw (l) -- (2em,18em) -- (37em,18em) -- (d21);
%\draw (l) -- (2em,18em) -- (28.5em,18em) -- (d121);
% V = X
\draw (b12) -- (10em,16em) -- (33em,16em) -- (d122);
% V = X = 1
%\draw (b11) -- (6em,17em) -- (37em,17em) -- (d21);
\draw (b11) -- (6em,17em) -- (28.5em,17em) -- (d121);
\draw (j) -- (-2em,19em) -- (40em,19em) -- (d221);
\draw (d222) -- (45em,12em) -- (47em,12em) -- (s);
\end{tikzpicture}
\end{sideways}

\section{Nijlpaarden}
\label{app:nijl}

% Nijlpaarden
\begin{sideways}
\begin{tikzpicture}
\node (j) at (-2em,0em) {$\overset{-}{np(1,2)}$};
\node (h) at (2em,0em) {$\overset{-}{np(2,3)}$};
\node (c) at (6em,0em) {$\overset{-}{np(3,4)}$};
\node (dn) at (10em,0em) {$\overset{-}{np(4,6)}$};
\node (zb) at (19em,0em) {$\overset{-}{\multimap}$};
\node (z1) at (15em,3em) {$\overset{+}{\textit{inf}(V,6,7,W)}$};
\node (z2) at (22em,3em) {$\overset{-}{\multimap}$};
\node (z21) at (18.5em,6em) {$\overset{+}{np(W',V)}$};
\node (z22) at (25em,6em) {$\overset{-}{\multimap}$};
\node (z221) at (22.5em,9em) {$\overset{+}{np(V',W')}$};
\node (z222) at (27.5em,9em) {$\overset{-}{s(V',W)}$};
\draw (zb) -- (z1);
\draw (zb) -- (z2);
\draw (z2) -- (z21);
\draw (z2) -- (z22);
\draw (z22) -- (z221);
\draw (z22) -- (z222);
\node (hb) at (35em,0em) {$\overset{-}{\multimap}$};
\node (h1) at (31em,3em) {$\overset{+}{\textit{inf}(X,Y,8,Z)}$};
\node (h2) at (39em,3em) {$\overset{-}{\multimap}$};
\node (h21) at (35.5em,6em) {$\overset{+}{np(X',X)}$};
\node (h22) at (42em,6em) {$\overset{-}{\textit{inf}(X',Y,7,Z)}$};
\draw (hb) -- (h1);
\draw (hb) -- (h2);
\draw (h2) -- (h21);
\draw (h2) -- (h22);
\node (vb) at (49em,0em) {$\overset{-}{\multimap}$};
\node (v1) at (46em,3em) {$\overset{+}{np(Y',Z')}$};
\node (v2) at (52em,3em) {$\overset{-}{\textit{inf}(Y',Z',8,9)}$};
\draw (vb) -- (v1);
\draw (vb) -- (v2);
\node (s) at (56em,0em) {$\overset{+}{s(1,9)}$};
\node (jl) at (-2em,-2em) {Jan};
\node (hl) at (2em,-2em) {Henk};
\node (cl) at (6em,-2em) {Cecilia};
\node (dl) at (10em,-2em) {dn};
\node (zl) at (19em,-2em) {zag};
\node (hel) at (35em,-2em) {helpen};
\node (vl) at (49em,-2em) {voeren};
\end{tikzpicture}
\end{sideways}

\begin{sideways}
\begin{tikzpicture}
\node (j) at (-2em,0em) {$\overset{-}{np(1,2)}$};
\node (h) at (2em,0em) {$\overset{-}{np(2,3)}$};
\node (c) at (6em,0em) {$\overset{-}{np(3,4)}$};
\node (dn) at (10em,0em) {$\overset{-}{np(4,6)}$};
\node (zb) at (19em,0em) {$\overset{-}{\multimap}$};
\node (z1) at (15em,3em) {$\overset{+}{\textit{inf}(V,6,7,W)}$};
\node (z2) at (22em,3em) {$\overset{-}{\multimap}$};
\node (z21) at (18.5em,6em) {$\overset{+}{np(W',V)}$};
\node (z22) at (25em,6em) {$\overset{-}{\multimap}$};
\node (z221) at (22.5em,9em) {$\overset{+}{np(V',W')}$};
\node (z222) at (27.5em,9em) {$\overset{-}{s(V',W)}$};
\draw (zb) -- (z1);
\draw (zb) -- (z2);
\draw (z2) -- (z21);
\draw (z2) -- (z22);
\draw (z22) -- (z221);
\draw (z22) -- (z222);
\node (hb) at (35em,0em) {$\overset{-}{\multimap}$};
\node (h1) at (31em,3em) {$\overset{+}{\textit{inf}(X,Y,8,9)}$};
\node (h2) at (39em,3em) {$\overset{-}{\multimap}$};
\node (h21) at (35.5em,6em) {$\overset{+}{np(X',X)}$};
\node (h22) at (42em,6em) {$\overset{-}{\textit{inf}(X',Y,7,9)}$};
\draw (hb) -- (h1);
\draw (hb) -- (h2);
\draw (h2) -- (h21);
\draw (h2) -- (h22);
\node (vb) at (49em,0em) {$\overset{-}{\multimap}$};
\node (v1) at (46em,3em) {$\overset{+}{np(X,Y)}$};
\node (v2) at (52em,3em) {$\overset{-}{\textit{inf}(X,Y,8,9)}$};
\draw (vb) -- (v1);
\draw (vb) -- (v2);
\node (s) at (56em,0em) {$\overset{+}{s(1,9)}$};
\node (jl) at (-2em,-2em) {Jan};
\node (hl) at (2em,-2em) {Henk};
\node (cl) at (6em,-2em) {Cecilia};
\node (dl) at (10em,-2em) {dn};
\node (zl) at (19em,-2em) {zag};
\node (hel) at (35em,-2em) {helpen};
\node (vl) at (49em,-2em) {voeren};
% Y' = X, Z' = Y, Z = 9
\draw (v2) -- (52em,10em) -- (31em,10em) -- (h1);
\end{tikzpicture}
\end{sideways}

\begin{sideways}
\begin{tikzpicture}
\node (j) at (-2em,0em) {$\overset{-}{np(1,2)}$};
\node (h) at (2em,0em) {$\overset{-}{np(2,3)}$};
\node (c) at (6em,0em) {$\overset{-}{np(3,4)}$};
\node (dn) at (10em,0em) {$\overset{-}{np(4,6)}$};
\node (zb) at (19em,0em) {$\overset{-}{\multimap}$};
\node (z1) at (15em,3em) {$\overset{+}{\textit{inf}(V,6,7,9)}$};
\node (z2) at (22em,3em) {$\overset{-}{\multimap}$};
\node (z21) at (18.5em,6em) {$\overset{+}{np(W',V)}$};
\node (z22) at (25em,6em) {$\overset{-}{\multimap}$};
\node (z221) at (22.5em,9em) {$\overset{+}{np(V',W')}$};
\node (z222) at (27.5em,9em) {$\overset{-}{s(V',9)}$};
\draw (zb) -- (z1);
\draw (zb) -- (z2);
\draw (z2) -- (z21);
\draw (z2) -- (z22);
\draw (z22) -- (z221);
\draw (z22) -- (z222);
\node (hb) at (35em,0em) {$\overset{-}{\multimap}$};
\node (h1) at (31em,3em) {$\overset{+}{\textit{inf}(X,6,8,9)}$};
\node (h2) at (39em,3em) {$\overset{-}{\multimap}$};
\node (h21) at (35.5em,6em) {$\overset{+}{np(V,X)}$};
\node (h22) at (42em,6em) {$\overset{-}{\textit{inf}(V,6,7,9)}$};
\draw (hb) -- (h1);
\draw (hb) -- (h2);
\draw (h2) -- (h21);
\draw (h2) -- (h22);
\node (vb) at (49em,0em) {$\overset{-}{\multimap}$};
\node (v1) at (46em,3em) {$\overset{+}{np(X,6)}$};
\node (v2) at (52em,3em) {$\overset{-}{\textit{inf}(X,6,8,9)}$};
\draw (vb) -- (v1);
\draw (vb) -- (v2);
\node (s) at (56em,0em) {$\overset{+}{s(1,9)}$};
\node (jl) at (-2em,-2em) {Jan};
\node (hl) at (2em,-2em) {Henk};
\node (cl) at (6em,-2em) {Cecilia};
\node (dl) at (10em,-2em) {dn};
\node (zl) at (19em,-2em) {zag};
\node (hel) at (35em,-2em) {helpen};
\node (vl) at (49em,-2em) {voeren};
% Y' = X, Z' = Y, Z = 9
\draw (v2) -- (52em,10em) -- (31em,10em) -- (h1);
% X' = V, Y = 6, W = 9
\draw (h22) -- (42em,11em) -- (15em,11em) -- (z1);
\end{tikzpicture}
\end{sideways}

\begin{sideways}
\begin{tikzpicture}
\node (j) at (-2em,0em) {$\overset{-}{np(1,2)}$};
\node (h) at (2em,0em) {$\overset{-}{np(2,3)}$};
\node (c) at (6em,0em) {$\overset{-}{np(3,4)}$};
\node (dn) at (10em,0em) {$\overset{-}{np(4,6)}$};
\node (zb) at (19em,0em) {$\overset{-}{\multimap}$};
\node (z1) at (15em,3em) {$\overset{+}{\textit{inf}(V,6,7,9)}$};
\node (z2) at (22em,3em) {$\overset{-}{\multimap}$};
\node (z21) at (18.5em,6em) {$\overset{+}{np(W',V)}$};
\node (z22) at (25em,6em) {$\overset{-}{\multimap}$};
\node (z221) at (22.5em,9em) {$\overset{+}{np(V',W')}$};
\node (z222) at (27.5em,9em) {$\overset{-}{s(V',9)}$};
\draw (zb) -- (z1);
\draw (zb) -- (z2);
\draw (z2) -- (z21);
\draw (z2) -- (z22);
\draw (z22) -- (z221);
\draw (z22) -- (z222);
\node (hb) at (35em,0em) {$\overset{-}{\multimap}$};
\node (h1) at (31em,3em) {$\overset{+}{\textit{inf}(4,6,8,9)}$};
\node (h2) at (39em,3em) {$\overset{-}{\multimap}$};
\node (h21) at (35.5em,6em) {$\overset{+}{np(V,4)}$};
\node (h22) at (42em,6em) {$\overset{-}{\textit{inf}(V,6,7,9)}$};
\draw (hb) -- (h1);
\draw (hb) -- (h2);
\draw (h2) -- (h21);
\draw (h2) -- (h22);
\node (vb) at (49em,0em) {$\overset{-}{\multimap}$};
\node (v1) at (46em,3em) {$\overset{+}{np(4,6)}$};
\node (v2) at (52em,3em) {$\overset{-}{\textit{inf}(4,6,8,9)}$};
\draw (vb) -- (v1);
\draw (vb) -- (v2);
\node (s) at (56em,0em) {$\overset{+}{s(1,9)}$};
\node (jl) at (-2em,-2em) {Jan};
\node (hl) at (2em,-2em) {Henk};
\node (cl) at (6em,-2em) {Cecilia};
\node (dl) at (10em,-2em) {dn};
\node (zl) at (19em,-2em) {zag};
\node (hel) at (35em,-2em) {helpen};
\node (vl) at (49em,-2em) {voeren};
% Y' = X, Z' = Y, Z = 9
\draw (v2) -- (52em,10em) -- (31em,10em) -- (h1);
% X' = V, Y = 6, W = 9
\draw (h22) -- (42em,11em) -- (15em,11em) -- (z1);
% X = 4
\draw (v1) -- (46em,12em) -- (10em,12em) -- (dn);
\end{tikzpicture}
\end{sideways}

\begin{sideways}
\begin{tikzpicture}
\node (j) at (-2em,0em) {$\overset{-}{np(1,2)}$};
\node (h) at (2em,0em) {$\overset{-}{np(2,3)}$};
\node (c) at (6em,0em) {$\overset{-}{np(3,4)}$};
\node (dn) at (10em,0em) {$\overset{-}{np(4,6)}$};
\node (zb) at (19em,0em) {$\overset{-}{\multimap}$};
\node (z1) at (15em,3em) {$\overset{+}{\textit{inf}(3,6,7,9)}$};
\node (z2) at (22em,3em) {$\overset{-}{\multimap}$};
\node (z21) at (18.5em,6em) {$\overset{+}{np(W',3)}$};
\node (z22) at (25em,6em) {$\overset{-}{\multimap}$};
\node (z221) at (22.5em,9em) {$\overset{+}{np(V',W')}$};
\node (z222) at (27.5em,9em) {$\overset{-}{s(V',9)}$};
\draw (zb) -- (z1);
\draw (zb) -- (z2);
\draw (z2) -- (z21);
\draw (z2) -- (z22);
\draw (z22) -- (z221);
\draw (z22) -- (z222);
\node (hb) at (35em,0em) {$\overset{-}{\multimap}$};
\node (h1) at (31em,3em) {$\overset{+}{\textit{inf}(4,6,8,9)}$};
\node (h2) at (39em,3em) {$\overset{-}{\multimap}$};
\node (h21) at (35.5em,6em) {$\overset{+}{np(3,4)}$};
\node (h22) at (42em,6em) {$\overset{-}{\textit{inf}(3,6,7,9)}$};
\draw (hb) -- (h1);
\draw (hb) -- (h2);
\draw (h2) -- (h21);
\draw (h2) -- (h22);
\node (vb) at (49em,0em) {$\overset{-}{\multimap}$};
\node (v1) at (46em,3em) {$\overset{+}{np(4,6)}$};
\node (v2) at (52em,3em) {$\overset{-}{\textit{inf}(4,6,8,9)}$};
\draw (vb) -- (v1);
\draw (vb) -- (v2);
\node (s) at (56em,0em) {$\overset{+}{s(1,9)}$};
\node (jl) at (-2em,-2em) {Jan};
\node (hl) at (2em,-2em) {Henk};
\node (cl) at (6em,-2em) {Cecilia};
\node (dl) at (10em,-2em) {dn};
\node (zl) at (19em,-2em) {zag};
\node (hel) at (35em,-2em) {helpen};
\node (vl) at (49em,-2em) {voeren};
% Y' = X, Z' = Y, Z = 9
\draw (v2) -- (52em,10em) -- (31em,10em) -- (h1);
% X' = V, Y = 6, W = 9
\draw (h22) -- (42em,11em) -- (15em,11em) -- (z1);
% X = 4
\draw (v1) -- (46em,12em) -- (10em,12em) -- (dn);
% V = 3
\draw (h21) -- (35.5em,13em) -- (6em,13em) -- (c);
\end{tikzpicture}
\end{sideways}

\begin{sideways}
\begin{tikzpicture}
\node (j) at (-2em,0em) {$\overset{-}{np(1,2)}$};
\node (h) at (2em,0em) {$\overset{-}{np(2,3)}$};
\node (c) at (6em,0em) {$\overset{-}{np(3,4)}$};
\node (dn) at (10em,0em) {$\overset{-}{np(4,6)}$};
\node (zb) at (19em,0em) {$\overset{-}{\multimap}$};
\node (z1) at (15em,3em) {$\overset{+}{\textit{inf}(3,6,7,9)}$};
\node (z2) at (22em,3em) {$\overset{-}{\multimap}$};
\node (z21) at (18.5em,6em) {$\overset{+}{np(2,3)}$};
\node (z22) at (25em,6em) {$\overset{-}{\multimap}$};
\node (z221) at (22.5em,9em) {$\overset{+}{np(V',2)}$};
\node (z222) at (27.5em,9em) {$\overset{-}{s(V',9)}$};
\draw (zb) -- (z1);
\draw (zb) -- (z2);
\draw (z2) -- (z21);
\draw (z2) -- (z22);
\draw (z22) -- (z221);
\draw (z22) -- (z222);
\node (hb) at (35em,0em) {$\overset{-}{\multimap}$};
\node (h1) at (31em,3em) {$\overset{+}{\textit{inf}(4,6,8,9)}$};
\node (h2) at (39em,3em) {$\overset{-}{\multimap}$};
\node (h21) at (35.5em,6em) {$\overset{+}{np(3,4)}$};
\node (h22) at (42em,6em) {$\overset{-}{\textit{inf}(3,6,7,9)}$};
\draw (hb) -- (h1);
\draw (hb) -- (h2);
\draw (h2) -- (h21);
\draw (h2) -- (h22);
\node (vb) at (49em,0em) {$\overset{-}{\multimap}$};
\node (v1) at (46em,3em) {$\overset{+}{np(4,6)}$};
\node (v2) at (52em,3em) {$\overset{-}{\textit{inf}(4,6,8,9)}$};
\draw (vb) -- (v1);
\draw (vb) -- (v2);
\node (s) at (56em,0em) {$\overset{+}{s(1,9)}$};
\node (jl) at (-2em,-2em) {Jan};
\node (hl) at (2em,-2em) {Henk};
\node (cl) at (6em,-2em) {Cecilia};
\node (dl) at (10em,-2em) {dn};
\node (zl) at (19em,-2em) {zag};
\node (hel) at (35em,-2em) {helpen};
\node (vl) at (49em,-2em) {voeren};
% Y' = X, Z' = Y, Z = 9
\draw (v2) -- (52em,10em) -- (31em,10em) -- (h1);
% X' = V, Y = 6, W = 9
\draw (h22) -- (42em,11em) -- (15em,11em) -- (z1);
% X = 4
\draw (v1) -- (46em,12em) -- (10em,12em) -- (dn);
% V = 3
\draw (h21) -- (35.5em,13em) -- (6em,13em) -- (c);
% W' = 2
\draw (z21) -- (18.5em,14em) -- (2em,14em) -- (h);
\end{tikzpicture}
\end{sideways}

\begin{sideways}
\begin{tikzpicture}
\node (j) at (-2em,0em) {$\overset{-}{np(1,2)}$};
\node (h) at (2em,0em) {$\overset{-}{np(2,3)}$};
\node (c) at (6em,0em) {$\overset{-}{np(3,4)}$};
\node (dn) at (10em,0em) {$\overset{-}{np(4,6)}$};
\node (zb) at (19em,0em) {$\overset{-}{\multimap}$};
\node (z1) at (15em,3em) {$\overset{+}{\textit{inf}(3,6,7,9)}$};
\node (z2) at (22em,3em) {$\overset{-}{\multimap}$};
\node (z21) at (18.5em,6em) {$\overset{+}{np(2,3)}$};
\node (z22) at (25em,6em) {$\overset{-}{\multimap}$};
\node (z221) at (22.5em,9em) {$\overset{+}{np(1,2)}$};
\node (z222) at (27.5em,9em) {$\overset{-}{s(1,9)}$};
\draw (zb) -- (z1);
\draw (zb) -- (z2);
\draw (z2) -- (z21);
\draw (z2) -- (z22);
\draw (z22) -- (z221);
\draw (z22) -- (z222);
\node (hb) at (35em,0em) {$\overset{-}{\multimap}$};
\node (h1) at (31em,3em) {$\overset{+}{\textit{inf}(4,6,8,9)}$};
\node (h2) at (39em,3em) {$\overset{-}{\multimap}$};
\node (h21) at (35.5em,6em) {$\overset{+}{np(3,4)}$};
\node (h22) at (42em,6em) {$\overset{-}{\textit{inf}(3,6,7,9)}$};
\draw (hb) -- (h1);
\draw (hb) -- (h2);
\draw (h2) -- (h21);
\draw (h2) -- (h22);
\node (vb) at (49em,0em) {$\overset{-}{\multimap}$};
\node (v1) at (46em,3em) {$\overset{+}{np(4,6)}$};
\node (v2) at (52em,3em) {$\overset{-}{\textit{inf}(4,6,8,9)}$};
\draw (vb) -- (v1);
\draw (vb) -- (v2);
\node (s) at (56em,0em) {$\overset{+}{s(1,9)}$};
\node (jl) at (-2em,-2em) {Jan};
\node (hl) at (2em,-2em) {Henk};
\node (cl) at (6em,-2em) {Cecilia};
\node (dl) at (10em,-2em) {dn};
\node (zl) at (19em,-2em) {zag};
\node (hel) at (35em,-2em) {helpen};
\node (vl) at (49em,-2em) {voeren};
% Y' = X, Z' = Y, Z = 9
\draw (v2) -- (52em,10em) -- (31em,10em) -- (h1);
% X' = V, Y = 6, W = 9
\draw (h22) -- (42em,11em) -- (15em,11em) -- (z1);
% X = 4
\draw (v1) -- (46em,12em) -- (10em,12em) -- (dn);
% V = 3
\draw (h21) -- (35.5em,13em) -- (6em,13em) -- (c);
% W' = 2
\draw (z21) -- (18.5em,14em) -- (2em,14em) -- (h);
% V' = 1
\draw (z221) -- (22.5em,15em) -- (-2em,15em) -- (j);
\draw (z222) -- (27.5em,14em) -- (56em,14em) -- (s);
\end{tikzpicture}
\end{sideways}

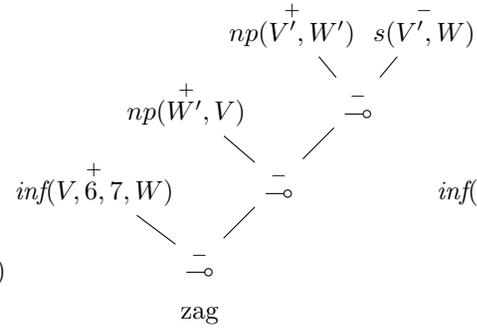
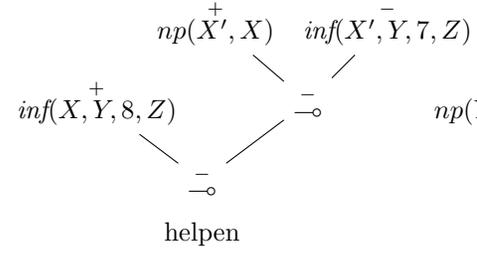
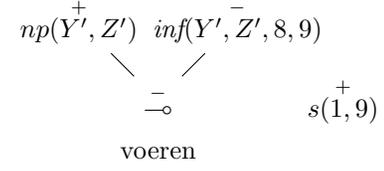
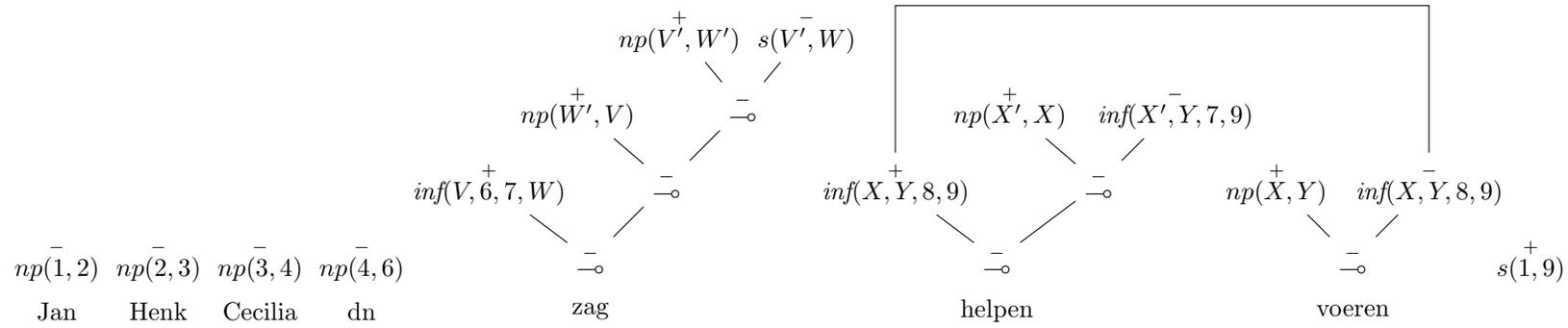
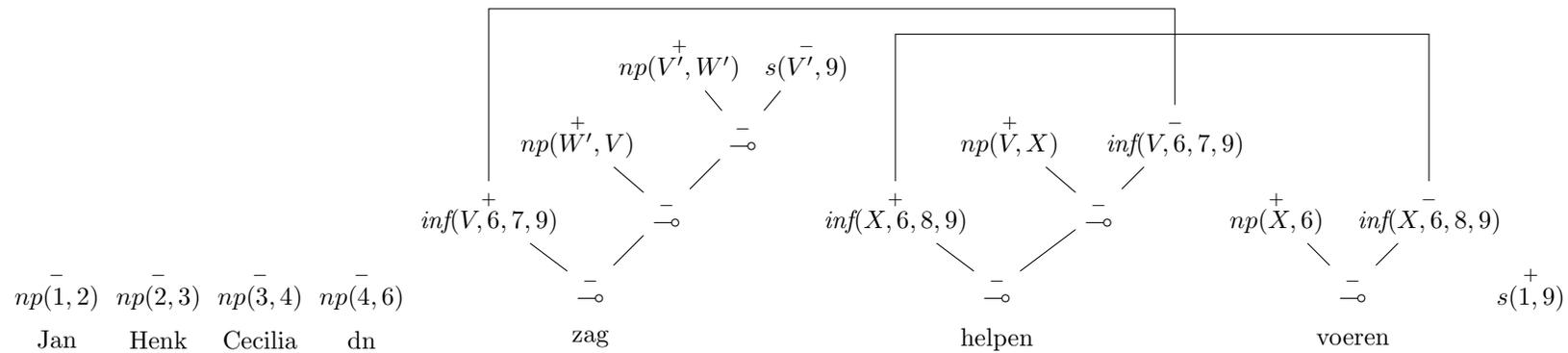
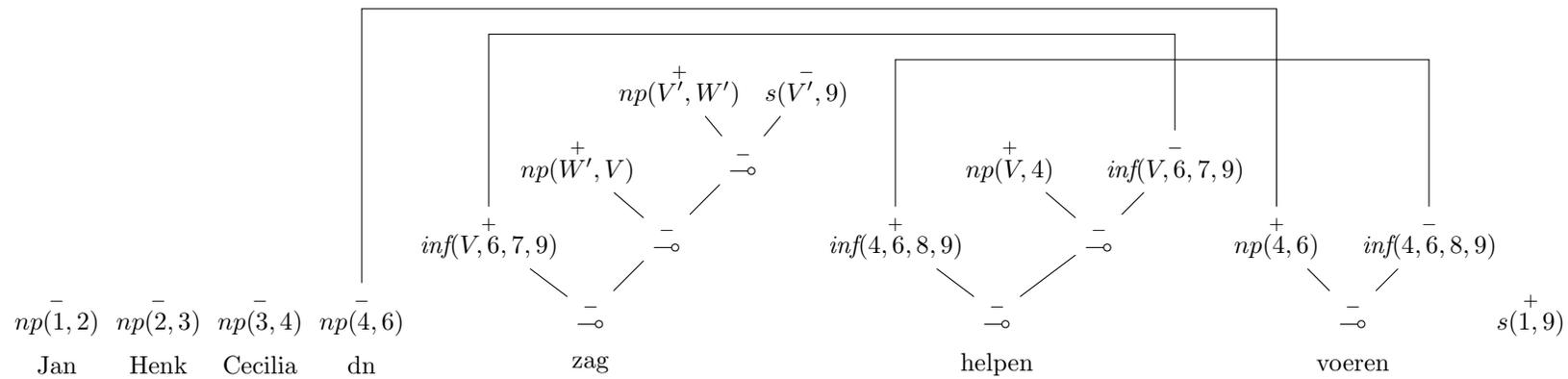
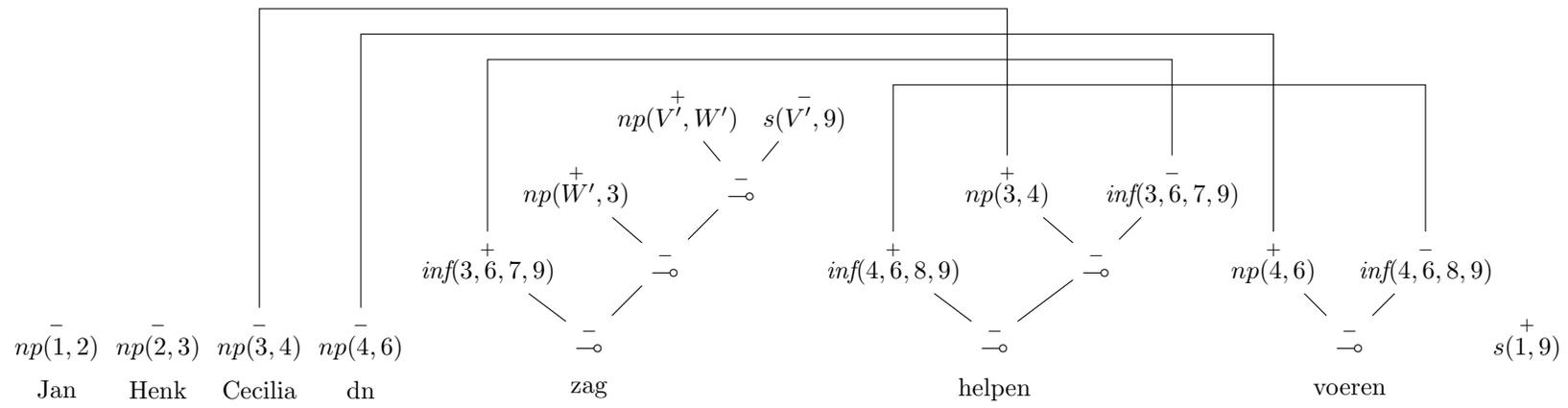
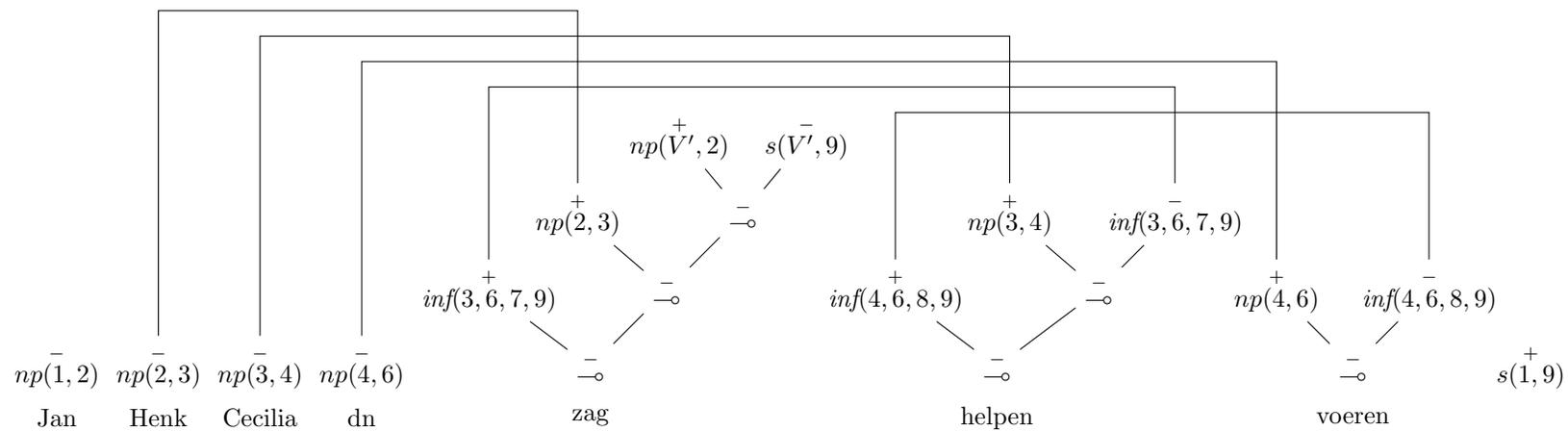
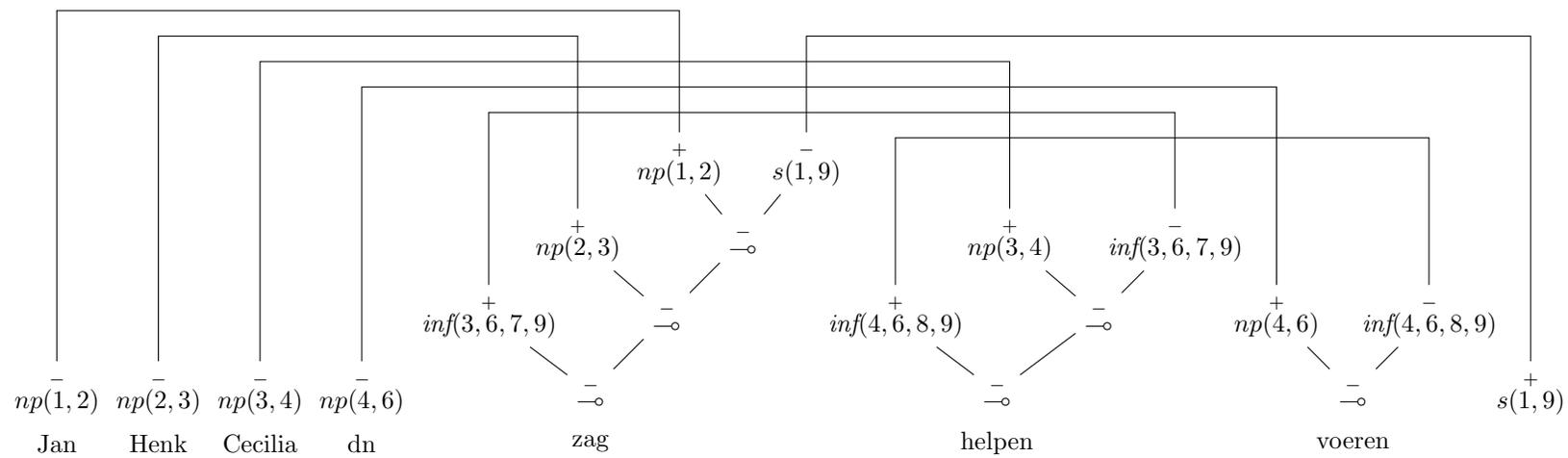
\end{document}